\documentclass[article,ij4uq]{ij4uq}              

\usepackage[hang]{footmisc}
\fancypagestyle{plain}{%
  \fancyhf{}
  \fancyhead[R]{\small {\it International Journal for Uncertainty Quantification}, x(x): \thepage--\pageref{LastPage} (\myyear\today)}
  \fancyfoot[R]{\small\bf\thepage }
  \fancyfoot[L]{\fottitle}
  }
\renewcommand{\dmyy}{17}
\renewcommand{\myyear}{2019}
\renewcommand{\today}{}

\usepackage{booktabs}       
\usepackage{amsfonts}       
\usepackage{nicefrac}       
\usepackage{microtype}      
\usepackage{lipsum,hyperref,subcaption}
\usepackage{tikz,pgfplots}
\pgfplotsset{compat=1.14}
\usepackage[opta]{optional} 

\usepgfplotslibrary{fillbetween}
\usepackage[outline]{contour}
\contournumber{64}
\contourlength{.12em}
\def\layersep{2.5cm}
\tikzstyle{int}=[draw, fill=blue!20, minimum size=2em]
\tikzstyle{init} = [pin edge={to-,thin,black}]

\usetikzlibrary{shapes.misc, positioning}

\usepackage{algorithmic}
\usepackage{multirow}

\makeatletter
\def\BState{\State\hskip-\ALG@thistlm}
\makeatother

\usepackage{amsmath,amssymb}
\usepackage[sort&compress]{natbib}

\newcommand{\D}{\mathcal{D}}
\newcommand{\Dh}{\mathcal{D}_h}
\newcommand{\Dl}{\mathcal{D}_l}

\newcommand{\gp}{\mathcal{GP}}

\newcommand{\mm}{\mathbf{m}}
\newcommand{\Ntr}{{N}}

\newcommand{\ppm}{\mathbf{p}}
\newcommand{\Km}{\mathbf{K}}
\newcommand{\km}{\mathbf{k}}

\newcommand{\vm}{\mathbf{v}}
\newcommand{\Xm}{\mathbf{X}}

\newcommand{\Wm}{\mathbf{W}}
\newcommand{\xm}{\mathbf{x}}

\newcommand{\ym}{\mathbf{y}}

\newcommand{\betaa}{\boldsymbol{\beta}}
\newcommand{\pinn}{\widehat{\mathcal{M}}}

\begin{document}

\volume{Volume x, Issue x, \myyear\today}
\title{On Transfer Learning of Neural Networks using Bi-fidelity Data for Uncertainty Propagation}
\titlehead{Transfer Learning for Uncertainty Propagation}
\authorhead{S. De, J. Britton, M. Reynolds, R. Skinner, K. Jansen, \& A. Doostan}
\author[1]{Subhayan De}
\author[2]{Jolene Britton}
\author[3]{Matthew Reynolds}
\author[1]{Ryan Skinner}
\author[1]{Kenneth Jansen}
\corrauthor[1]{Alireza Doostan}
\corremail{doostan@colorado.edu}
\corraddress{University of Colorado, Boulder, CO 80309}
\address[1]{University of Colorado, Boulder, CO 80309}
\address[2]{University of California, Riverside, CA 92521}
\address[3]{National Renewable Energy Laboratory, Golden, CO 80401}

\dataO{mm/dd/yyyy}
\dataF{mm/dd/yyyy}

\abstract{Due to their high degree of expressiveness, neural networks have recently been used as surrogate models for mapping inputs of an engineering system to outputs of interest. Once trained, neural networks are computationally inexpensive to evaluate and remove the need for repeated evaluations of computationally expensive models in uncertainty quantification applications. However, given the highly parameterized construction of neural networks, especially deep neural networks, accurate training often requires large amounts of simulation data that may not be available in the case of computationally expensive systems. In this paper, to alleviate this issue for uncertainty propagation, we explore the application of transfer learning techniques using training data generated from both high-and low-fidelity models. We explore two strategies for coupling these two datasets during the training procedure, namely, the standard transfer learning and the bi-fidelity weighted learning. In the former approach, a neural network model mapping the inputs to the outputs of interest is trained based on the low-fidelity data. The high-fidelity data is then used to adapt the parameters of the upper layer(s) of the low-fidelity network, or train a simpler neural network to map the output of the low-fidelity network to that of the high-fidelity model. In the latter approach, the entire low-fidelity network parameters are updated using data generated via a Gaussian process model trained with a small high-fidelity dataset. The parameter updates are performed via a variant of stochastic gradient descent with learning rates given by the Gaussian process model. Using three numerical examples, we illustrate the utility of these bi-fidelity transfer learning methods where we focus on accuracy improvement achieved by transfer learning over standard training approaches. 
}


\maketitle

\section{Introduction}\label{sec:intro}

Neural networks build on imitating the biological neurons and are useful for learning nonlinear and 
complex functions \cite{gurney2014introduction,goodfellow2016deep}. With the recent advances in computing resources and software, it is now possible to train large neural networks that accurately describe the response of physical systems.
Building surrogate models for mechanics problems using neural networks offers several distinct benefits. In particular, the past few years have seen a proliferation of open source software tools for building and training neural networks. Examples of widely used software packages include Tensorflow \cite{abadi2016tensorflow}, Theano \cite{bergstra2010theano}, Caffe \cite{jia2014caffe}, and PyTorch \cite{paszke2017automatic}. These tools supply users with efficient software abstractions for building and training neural networks, resulting in accelerated workflows that produce expressive surrogate models in a small amount of time. Furthermore, these software packages enjoy large user community support outside of applications in science and engineering. 
Another benefit of building surrogate models using neural networks is the ability to leverage specialized hardware architectures that enable faster training of models. For example, the software packages listed above all contain some consideration toward training neural networks using GPUs. 
More importantly, large investments in high performance computing resources specifically tailored for artificial intelligence (AI) applications are becoming more commonplace. For example, at the time of publication of this article, the first two exascale computers planned for delivery to the Department of Energy will include some degree of optimization for AI applications \cite{doeintel2019,doeamd2018}. Also, the current top two spots on the Top500 list \cite{top500list2019} are occupied by high-performance computing systems built by IBM that include NVIDIA Volta GV100 GPUs. 
Finally, neural networks have the potential to have a high level of expressiveness. The composite function model has been shown to have a universal approximation property given the use of differentiable activation functions and a single layer with a large number of neurons \cite{cybenko1989approximation,hornik1989multilayer,hornik1990universal,opschoor2019deep}. 

Despite the above said benefits of using neural networks as surrogate models, there are some limitations. One such 
limitation is the lack of clarity involving the application of physical constraints or symmetries of the models. This limitation is well-known and is currently an area of focus in the Scientific Machine Learning (SciML) community \cite{baker2019workshop}. 
However, the most limiting drawback of using neural networks as surrogate models is perhaps the substantial---at times prohibitive for scientific applications---data requirements for training neural networks. 
Datasets for scientific and engineering applications are often limited in size due to the costs associated with the physical experiments or the computational complexity of high-fidelity simulations. A potential cure for this lack of data volume is the integration of data from models or experiments of different fidelities into the neural network training process. 

Multiple models in computational mechanics are often available to describe a physical phenomenon ({\textit{e.g.}}, turbulence) \cite{fernandez2016review,peherstorfer2018survey}.  
Among them, some models, known as the \textit{high-fidelity} models, require more computational resources but provide higher levels of accuracy. These models can be used to generate accurate data for training neural networks. However, it may be computationally expensive or even infeasible to do so, even when running in parallel. 
On the other hand, models with generally low levels of accuracy often (not always) require small computational budget and can be used to generate a large dataset. These models are denoted as \textit{low-fidelity} models. As an example, a coarse finite element discretization of a differential equations may give rise to a low-fidelity model as compared to a fine discretization of the same problem that achieves a desired accuracy.  Other instances of low-fidelity models include reduced order models \cite{sirovich1987turbulence,berkooz1993proper,benner2015survey} or regression models \cite{rasmussen2004gaussian,martin2005use,forrester2009recent}. 

In a multi-fidelity approach, the availability of these different fidelities of models are exploited \cite{fernandez2016review,peherstorfer2018survey} to achieve a high level of accuracy in the response prediction. In this approach, the smaller computational cost of the low-fidelity models are utilized and a smaller number of high-fidelity model evaluations are used to {\it correct} the error introduced through using many low-fidelity model evaluations. Related to the topic of this study, multi-fidelity methods have also shown to lead to effective uncertainty quantification strategies. In \cite{koutsourelakis2009accurate}, a Bayesian model with the sequential Monte Carlo algorithm is implemented to establish a relation between inaccurate low-fidelity models and high-fidelity models. Li and Xiu \cite{li2010evaluation}, Li \textit{et al.} \cite{li2011efficient}, and Peherstorfer \textit{et al.} \cite{peherstorfer2017combining} used a combination of low- and high-fidelity models to evaluate probability of failure. Additive and multiplicative corrections of low-fidelity models via small number of high-fidelity samples have been developed for Gaussian process regression \cite{Kennedy00,Qian06,Kleiber13,LeGratiet14} (a.k.a. co-kriging) and polynomial chaos expansions \cite{Eldred09,ng2012multifidelity, Padron16}. A different class of methods rely on low-fidelity data to generate a low-rank approximation of high-fidelity quantities of interest, see, {\textit{e.g.}, \cite{Doostan07,Ghanem07a,narayan2014stochastic,zhu2014computational,doostan2016bi,hampton2018practical,fairbanks2018bi,skinner2019reduced}

In the domain of neural networks, transfer learning \cite{pan2010survey,weiss2016survey} attempts to use the knowledge gained from solving one problem to solve another related problem, where accurate or labeled data may be missing or limited. 
For example, when labeled data for two related but different classification or regression tasks are available, knowledge gained from training a neural network for one of the problems can be used to help training for the other one. This scenario is known as the \textit{inductive transfer learning} \cite{pan2010survey}. Similarly, in another scenario, transfer learning can be implemented for two unrelated tasks. This application of transfer learning is known as \textit{transductive transfer learning}. 
These different types of transfer learning have seen many applications in machine learning domain, including text classification \cite{li2012cross}, image classification \cite{zhu2011heterogeneous}, natural language processing \cite{collobert2008unified}, and face recognition \cite{kan2014domain}. Alongside, there are applications of transfer learning in tuning global climate models \cite{ma2015transfer}, disease prediction \cite{ogoe2015knowledge}, genome biology \cite{widmer2012multitask} and so on. 
For a detailed list of applications we refer the interested reader to \cite{weiss2016survey} and the references therein. Apart form these standard transfer learning scenarios, 
fidelity-weighted learning \cite{dehghani2017neural,dehghani2018fidelity} is an example of \textit{machines-teaching-machines}  or \textit{learning using privileged information} paradigms \cite{hinton2006fast,vapnik2009new,vapnik2015learning}, and can be thought of a transfer learning scenario. In this case, a neural network, named as the \textit{student}, is trained using easily available poorly labeled data. On the other hand, a Gaussian process is constructed using high-quality labeled data and known as the \textit{teacher}. The teacher's knowledge is then used to fine-tune the \textit{student} network. Dehghani \textit{et al}. \cite{dehghani2018fidelity} showed that in presence of large enough high-quality dataset the teacher can be used to improve the performance of the student network.

Generally speaking, transfer learning is demonstrated to provide the following advantages over standard learning, \cite{torrey2010transfer,pan2010survey}: (i) smaller initial training error, (ii) faster convergence of the optimization scheme used for training, and (iii) similar (or smaller) validation error using smaller datasets. In this work, we are particularly interested in the last advantage as in many scientific applications, training datasets are of limited size. However, we note that depending on the required accuracy and the availability of high-fidelity data, this advantage may diminish when a sufficient number of high fidelity samples are used in the training of the standard neural network. Therefore, the primary utility of transfer learning techniques is in scenarios where limited high-fidelity data is available (see Fig. \ref{fig:bftl_concept}). In addition, when the two dataset used in transfer learning are considerably dissimilar, transfer learning may lead to less accurate predictions, a phenomenon referred to as \textit{negative transfer} \cite{rosenstein2005transfer,ge2014handling,wang2019characterizing} and depicted in Fig. \ref{fig:bftl_concept}.

\begin{figure}[!htb]
    \centering
    \begin{tikzpicture}
     \begin{axis}[thick,smooth,no markers,axis lines=none, xtick=\empty, ytick=\empty]
        \addplot [very thick, blue, smooth, name path = A,domain = 0:3] coordinates {(0,.2) (1.6,-1) (3.0,-1.2)};
        \addplot [very thick, red, dashdotted, smooth, name path = B,domain = 0:3] coordinates {(0,0) (0.7,-1.15) (3,-1.25)};
        \addplot [very thick, -latex] coordinates {(0,-1.35) (0,0.5)};
        \addplot [very thick, -latex] coordinates {(0,-1.35) (3.35,-1.35)};
        \addplot [thin, name path = C] coordinates {(0,-1.35) (3,-1.35)};
            \addplot[blue!40] fill between[of=A and B];
            \addplot[red!20] fill between[of=B and C];
        \addplot [very thick, teal!80!black, dashed, smooth, name path = B,domain = 0:3] coordinates {(0,0) (0.7,-0.95) (3,-1)};
        
    \draw[very thick, red, fill = red!60] (axis cs: 1.65,-1.025) circle [radius=0.025];
\end{axis}
\node at (3.35, 0.1)   (a) { Equivalent number of high-fidelity samples};
\node[rotate = 90] at (0.25, 2.5)   (b) { Prediction Error};


\draw[thick,fill=green!5] (3,3.25) rectangle (5.7,4.6);
\draw[very thick,blue] (3.25,4.35) -- (3.75,4.35);
\draw[very thick,red, dashdotted] (3.25,3.95) -- (3.75,3.95);
\draw[very thick,teal!80!black, dashed] (3.25,3.55) -- (3.75,3.55);

\node[draw=none,right] at (3.75,4.35) () {\small Without TL};
\node[draw=none,right] at (3.75,3.95) () {\small With TL};
\node[draw=none,right] at (3.75,3.55) () {\small Negative TL};

\draw[thick, latex-latex] (2,2.725) -- (2,0.85);
\node[draw = none,right] at (2,2) () {\contour{white}{\small TL advantage}};
\draw[thick, latex-latex] (5.65,1.4) -- (5.65,0.85);
\node[draw = none,right] at (5.7,1.125) () {{\small TL disadvantage}};
\draw[thick, latex-] (3.45,1.4) -- (4.8,1.8);
\node[draw = none,right] at (4.8,1.8) () {{\small Crossover}};
\end{tikzpicture}
    \caption{When the number of high-fidelity data is not adequate, the goal of transfer learning (TL) is to provide a means for robust and accurate neural network training, especially in small sample size regimes. As transfer learning relies on similar or approximate datasets (here of lower fidelity), the resulting network may lead to less accurate surrogates as compared to when the network is trained based only on large high-fidelity datasets. This so-called transfer learning disadvantage may not be necessarily eliminated by adding more high-fidelity samples.}
    \label{fig:bftl_concept}
\end{figure}
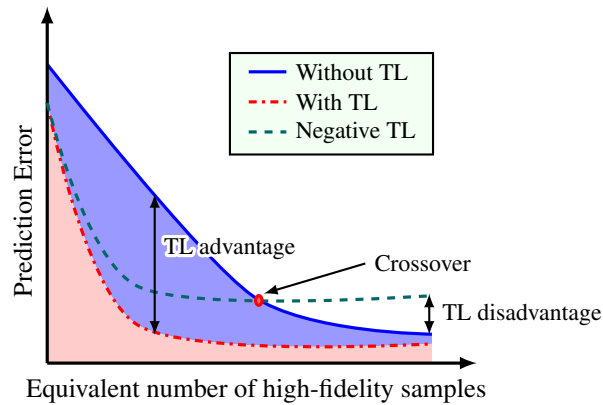


In the present study, we examine the utility of transfer learning in building neural network surrogates using a combination of low- and high-fidelity data. As the low-fidelity models can be simulated relatively cheaply, this allows access to a large quantity of training data without significant computational expense. Since the majority of the training data may be of lesser quality, we must couple the two types of data in a manner that the bi-fidelity predictions are still as accurate as possible. To this end, we use two transfer learning techniques, namely, the inductive transfer learning and the fidelity-weighted learning to efficiently train a neural network for uncertainty propagation. We use three numerical examples to show the usefulness of this approach. In the first example, bending of a composite beam with multiple circular holes on its web is considered, where the material properties and external load are uncertain. The low-fidelity model uses the Euler-Bernoulli beam equation whereas the high-fidelity model uses a finite element model. The second example considers a non-linear, multi-scale, multi-physics, Lithium ion battery model with 17 uncertain parameters, where the low- and high-fidelity models are, respectively, based on a coarse and fine discretization of the spatial domain.  The final example involves turbulent flow around an airfoil with parametric geometry and inflow angle-of-attack. The low-fidelity model solves the Reynolds averaged Navier-Stokes (RANS) equations using 9,000 elements in the finite element discretization whereas the high-fidelity model uses 241,000 elements. These three numerical examples show that the transfer learning approach can be efficiently used to train neural networks for uncertainty propagation in the presence of limited, high-fidelity training data.

Before proceeding, we remark that the techniques proposed in this paper specifically address the inclusion of bi-fidelity datasets into the neural network training process for developing surrogate models. The purpose of this work is not to claim the superiority of neural networks as surrogate models over other representations, {\textit{e.g.}, polynomial expansions or Gaussian processes. Comparing the performance of neural network surrogates with those based on well-established techniques in the literature is important, but beyond the scope of this work.

The rest of this paper is organized as follows. In Section \ref{sec:methodology}, we briefly discuss the components of a neural network and its training procedure. Following that, different aspects of transfer learning and fidelity-weighted learning for neural networks are introduced and the considered algorithms for using training bi-fidelity data are presented. In Section \ref{sec:ex}, three numerical examples are used to illustrate the efficacy of the two transfer learning techniques. Finally, we conclude the paper and mention future research directions.

\section{Methodology}
\label{sec:methodology}

In this section, after briefly introducing uncertainty propagation through a system and the working principles of neural networks, transfer learning methods using bi-fidelity training data are discussed. 

\subsection{Uncertainty Propagation}

Models, often in the form of partial differential equations, are used ubiquitously in order to describe the behavior of a physical system. The system's response or the quantity of interest  (QoI) $y$ -- here for simplicity assumed a scalar -- is often affected by the presence of uncertainty in its inputs $\xm\in\mathbb{R}^d$ describing, for instance, loading, boundary conditions, material properties, or the structure of the model itself. An instance of the utility of models is to relate the QoI $y$ to its inputs $\xm$, {\textit{i.e.}},
\begin{equation}\label{eq:model}
y=\mathcal{M}(\xm).
\end{equation}
The task of uncertainty propagation is to quantify the uncertainty in $y$ given the description of uncertainty in $\xm$. In probabilistic methods, the inputs $\xm$ are random variables with prescribed probability distributions. As such, the QoI $y$ is also a random variable and the goal of uncertainty propagation is to compute the statistics of $y$, \textit{e.g.}, mean and variance, or probability distribution function. One approach to uncertainty propagation, known as surrogate modeling, is to find an approximation to the map $\mathcal{M}$, denoted by $\widehat{\mathcal{M}}$, from samples of $\xm$ and $y$. Once $\widehat{\mathcal{M}}$ is constructed, the statistics of $y$ or the sensitivity of $y$ to the components of $\xm$ are estimated by simulating $\widehat{\mathcal{M}}$ -- instead of $\mathcal{M}$ -- which does not require evaluation of $\mathcal{M}$. 

When $\mathcal{M}$ is expensive to simulate, building an accurate surrogate $\widehat{\mathcal{M}}$ becomes computationally intensive, and an active area of research is to {\it minimize} the number of evaluations of $\mathcal{M}$ with little loss of accuracy for such a construction. In the present work, we rely on neural networks as a means to generate $\widehat{\mathcal{M}}$ and, as motivated in Section \ref{sec:intro}, bi-fidelity strategies to reduce the cost of building $\widehat{\mathcal{M}}$. We next provide a brief introduction to neural networks and their training.


%

%
%
\subsection{Neural Network Surrogates}

An artificial neural network, or simply neural network, mimics the behavior of neuron cells that process biological information. The recent advances in computing power have enabled the training of large neural networks to learn relationships formalized by \eqref{eq:model} assuming sufficient training data is available.
 The most commonly used neural network is known as the feed-forward neural network (FNN) or multilayer perceptron (MLP) \cite{goodfellow2016deep}. An FNN consists of a sequence of connected layers, including an input layer, one or more hidden layers, and an output layer (see Fig. \ref{fig:ffn_dnn}). Each of these hidden layers have multiple neurons which perform an affine transformation of output from the previous layer. Then an often nonlinear activation function is applied resulting in the neuron's output.
Mathematically, a neural network features a composite map between the inputs $\xm$ and the output $y$ given by
\begin{equation}\label{eq:nn}
\begin{split}
y &\approx \pinn\left(\xm;\{\Wm_i\}_{i=0}^{H}, \{\betaa_i\}_{i=0}^{H} \right)\\
 &:= \Wm_0\phi_{H}(\dots\phi_2(\Wm_2(\phi_1(\Wm_1\xm+\betaa_1)+\betaa_2)\dots)+\betaa_0,\\
\end{split}
\end{equation}
where $H$ is the number of hidden layers; $\{\Wm_i\}_{i=1}^{H}$ and $\{\betaa_i\}_{i=1}^{H}$ are, respectively, the unknown weight matrices and bias vectors for the $i$th hidden layer; $\Wm_0$ and $\betaa_0$ are -- with a slight abuse of notation -- the unknown weight vector and scalar bias for the output layer, respectively; and $\phi_i(\cdot)$ are (vector-valued) activation functions for the $i$th hidden layer. The goal of training the FNN model $\pinn$, or neural networks in general, is to determine the weights and biases  $\{\Wm_i\}_{i=0}^{H}$ ,$\{\betaa_i\}_{i=0}^{H}$,  so that $\pinn$ is close to $\mathcal{M}$ in some sense. 


\begin{figure}[!htb]
    \centering

\begin{tikzpicture}[shorten >=1pt,->,draw=black!50, node distance=\layersep,thick,latex-,scale=0.7]
    \tikzstyle{every pin edge}=[shorten <=1pt,thick]
    \tikzstyle{neuron}=[circle,fill=black,minimum size=14pt,inner sep=0pt,thick]
    \tikzstyle{input neuron}=[neuron, draw=black,fill=green!50!blue,];
    \tikzstyle{output neuron}=[neuron, draw=black, fill=blue!70,thick,-latex];
    \tikzstyle{hidden neuron}=[neuron, draw=black, fill=red!70,thick,-latex];
    \tikzstyle{annot} = [text width=4em, text centered]

    \foreach \name / \y in {1,...,2}
        \node[input neuron, thick,-latex,pin=left:$x_{\y}$] (I-\name) at (0,-\y-3) {};
    \foreach \name / \y in {3,...,3}
            \node[input neuron, thick,-latex,pin=left:$\vdots$] (I-\name) at (0,-\y-3) {};
    \foreach \name / \y in {4,...,4}
            \node[input neuron, thick,-latex,pin=left:$x_{d-1}$] (I-\name) at (0,-\y-3) {};    
    \foreach \name / \y in {5,...,5}
            \node[input neuron, thick,-latex,pin=left:$x_d$] (I-\name) at (0,-\y-3) {};    

    \foreach \name / \y in {1,...,9}
        \path[yshift=-1cm]
            node[hidden neuron] (H1-\name) at (\layersep,-\y cm) {};

      \foreach \name / \y in {1,...,9}
        \path[yshift=-1cm]
            node[hidden neuron] (H2-\name) at (2*\layersep,-\y cm) {};

     
    \node[output neuron,-latex,pin={[pin edge={-latex}]right:${y}$}, right of=H2-5] (O-1) {};
   
    \foreach \source in {1,...,5}
        \foreach \dest in {1,...,9}
            \path[thick,-latex] (I-\source) edge (H1-\dest);

  \foreach \source in {1,...,9}
        \foreach \dest in {1,...,9}
            \path[thick,-latex] (H1-\source) edge (H2-\dest);

    \foreach \source in {1,...,9}
       \foreach \dest in {1,...,1}
         \path[thick,-latex] (H2-\source) edge (O-\dest);

    \node[annot,above of=H1-1, node distance=1cm] (hl1) {Hidden layer 1};
    \node[annot,above of=H2-1, node distance=1cm] (hl2) {Hidden layer 2};
    \node[annot,above of=I-1, node distance=1cm] {Input layer};
    \node[annot,above of=O-1, node distance=1cm] {Output layer};

\end{tikzpicture}
\caption{Feed-forward neural network (FNN) architecture with two hidden layers.}
    \label{fig:ffn_dnn}
\end{figure}
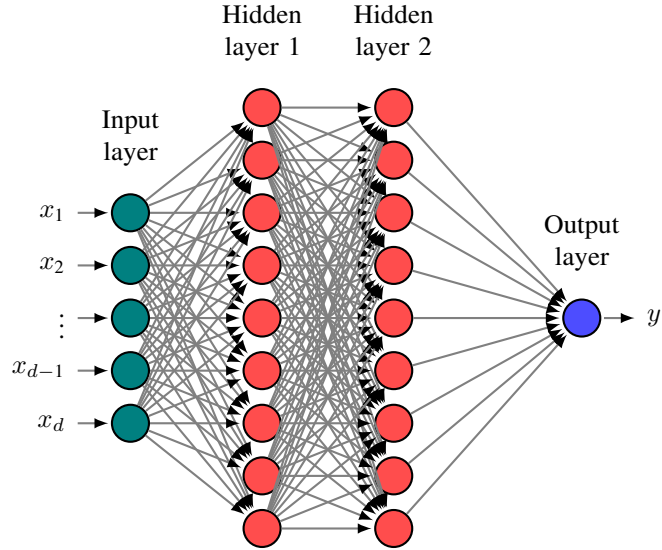

In another popular neural network architecture, dubbed residual network (ResNet), the output of a layer is directly added to the output of a subsequent layer, \textit{e.g.},  
\begin{equation}
\ym_i=\phi_i(\Wm_i\ym_{i-1}+\betaa_i)+\ym_{i-1},
\end{equation}
where here $\ym_{i-1}$ and $\ym_i$ are the outputs of the $(i-1)$th and $i$th layers, respectively. Stated differently, the $i$th layer models the residual between the $(i-1)$th and $i$th hidden layers \cite{he2016deep}, hence the name ResNet. Note that if $\ym_i$ and $\ym_{i-1}$ have different dimensions, a short-cut mapping as described in \cite{he2016deep} is needed. An illustration of the ResNet is shown in Fig. \ref{fig:resnet}. 

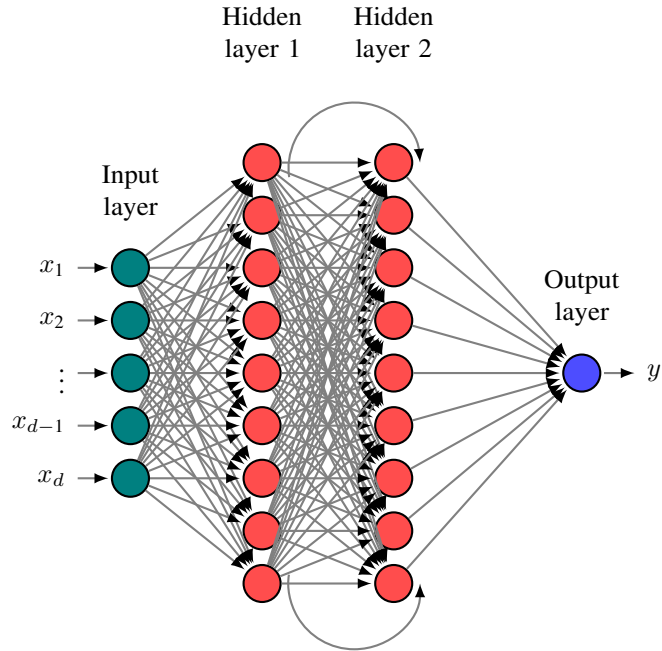
\begin{figure}
    \centering
\begin{tikzpicture}[shorten >=1pt,->,draw=black!50, node distance=\layersep,thick,latex-,scale=0.7]
    \tikzstyle{every pin edge}=[<-,shorten <=1pt,thick,latex-]
    \tikzstyle{neuron}=[circle,fill=black,minimum size=14pt,inner sep=0pt,thick,latex-]
    \tikzstyle{input neuron}=[neuron, draw=black,fill=green!50!blue,];
    \tikzstyle{output neuron}=[neuron, draw=black, fill=blue!70,thick,-latex];
    \tikzstyle{hidden neuron}=[neuron, draw=black, fill=red!70,thick,-latex];
    \tikzstyle{annot} = [text width=4em, text centered]

    \foreach \name / \y in {1,...,2}
        \node[input neuron, thick,-latex,pin=left:$x_{\y}$] (I-\name) at (0,-\y-3) {};
    \foreach \name / \y in {3,...,3}
            \node[input neuron, thick,-latex,pin=left:$\vdots$] (I-\name) at (0,-\y-3) {};
    \foreach \name / \y in {4,...,4}
            \node[input neuron, thick,-latex,pin=left:$x_{d-1}$] (I-\name) at (0,-\y-3) {};    
    \foreach \name / \y in {5,...,5}
            \node[input neuron, thick,-latex,pin=left:$x_{d}$] (I-\name) at (0,-\y-3) {};    

    \foreach \name / \y in {1,...,9}
        \path[yshift=-1cm]
            node[hidden neuron] (H1-\name) at (\layersep,-\y cm) {};

      \foreach \name / \y in {1,...,9}
        \path[yshift=-1cm]
            node[hidden neuron] (H2-\name) at (2*\layersep,-\y cm) {};

     
    \node[output neuron,-latex,pin={[pin edge={-latex}]right:${y}$}, right of=H2-5] (O-1) {};
   
    \foreach \source in {1,...,5}
        \foreach \dest in {1,...,9}
            \path[thick,-latex] (I-\source) edge (H1-\dest);

  \foreach \source in {1,...,9}
        \foreach \dest in {1,...,9}
            \path[thick,-latex] (H1-\source) edge (H2-\dest);

    \foreach \source in {1,...,9}
       \foreach \dest in {1,...,1}
         \path[thick,-latex] (H2-\source) edge (O-\dest);

    \node[annot,above of=H1-1, node distance=1.7cm] (hl1) {Hidden layer 1};
    \node[annot,above of=H2-1, node distance=1.7cm] (hl2) {Hidden layer 2};
    \node[annot,above of=I-1, node distance=1cm] {Input layer};
    \node[annot,above of=O-1, node distance=1cm] {Output layer};

    \draw[latex-,thick] (5.5,-2) arc (5:190:1.25);
    \draw[latex-,thick] (5.5,-10.) arc (0:-190:1.25);
\end{tikzpicture}
\caption{Residual neural network (ResNet) architecture with two hidden layers. The curved arrows illustrate skip connections for adding the output of a layer directly to that of another layer.}
    \label{fig:resnet}
\end{figure}

\subsubsection{Activation Functions}

There are many choices available for the activation function $\phi_i(\cdot)$ in \eqref{eq:nn}. A common choice is the rectified linear unit (ReLU), where the output is given by
\begin{equation}
\phi_{\!_\mathrm{ReLU}}(z) = \max(0,z),
\end{equation}
for an input $z$.
Figure \ref{fig:relu} shows the plot of an ReLU function and its derivative. Note that the derivative of the output vanishes for $z<0$. This creates difficulty during the training using gradient descent (see Section \ref{sec:training}) for negative inputs and is known as the \textit{dying ReLU} problem. To avoid this, we also investigate the use of another activation function, namely, exponential linear unit or ELU. In this case, the output is given by 
\begin{equation}
    \phi_{\!_\mathrm{ELU}}(z) = \begin{cases}
    z \qquad \qquad \quad~~~\! \text{for } z>0,\\
    \alpha (e^z-1) \qquad \text{for } z\leq0,\\
    \end{cases}
\end{equation}
where $\alpha$ is a positive parameter. Figure \ref{fig:elu} shows the difference between ELU and ReLU and their derivatives $z\leq0$. Note that there are other alternatives for activation functions, \textit{e.g.}, leaky ReLU, logistic sigmoid, hyperbolic tangent, etc. However, they are not discussed in this paper as we found the ELU function to produce the best results for the numerical examples of Section \ref{sec:ex}.


\begin{figure}[!htb]
\centering
\begin{tikzpicture}
    \begin{axis}[scale=0.75,
        domain=-3:5,x label style={at={(axis description cs:0.5,-0.1)},anchor=north},
            y label style={at={(axis description cs:-0.1,.5)},rotate=0,anchor=south},
            xlabel={$z$},
            ylabel={$\phi_{\!_\mathrm{ReLU}}(z)$}
        ]
        \addplot+[mark=none,blue,domain=-3:0,ultra thick] {0};
        \addplot+[mark=none,blue,domain=0:5,ultra thick] {x};
        \addplot+[mark=none,dashed] coordinates {(-6, 0) (4, 0)};
        \addplot+[mark=none,dashed] coordinates {(0, -1) (0, 5)};
    \end{axis}
\end{tikzpicture}
\hspace{.5cm}
\centering
\begin{tikzpicture}
    \begin{axis}[scale=0.75,
        domain=-3:5,x label style={at={(axis description cs:0.5,-0.1)},anchor=north},
            y label style={at={(axis description cs:-0.1,.5)},rotate=0,anchor=south},
            xlabel={$z$},
            ylabel={$\frac{d\phi_{\!_\mathrm{ReLU}}(z)}{dz}$}
        ]
        \addplot+[mark=none,blue,domain=-3:0,ultra thick] {0};
        \addplot+[mark=none,blue,domain=0:0.001,ultra thick] {1000*x};
        \addplot+[mark=none,dashed] coordinates {(-6, 0) (4, 0)};
        \addplot+[mark=none,blue,domain=0.001:5,ultra thick] {1};
    \end{axis}
\end{tikzpicture}
\caption{Rectified linear unit (ReLU) and its derivative.}\label{fig:relu}
\end{figure}
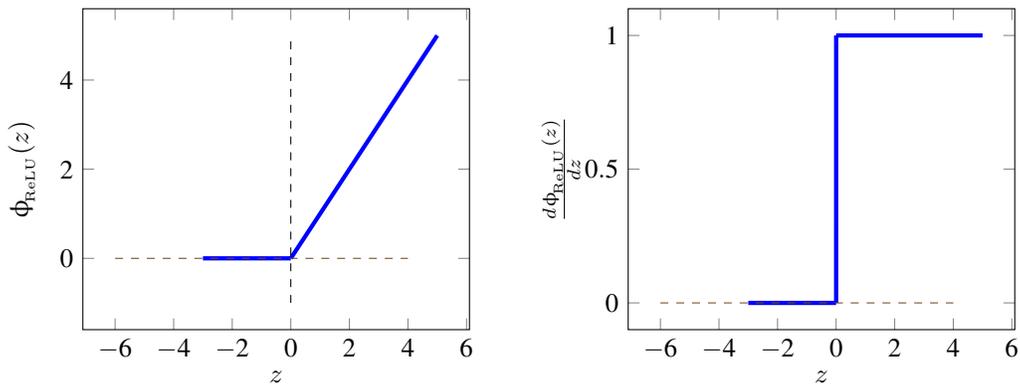

\begin{figure}[!htb]
\centering
\begin{tikzpicture}
    \begin{axis}[scale=0.75,
        domain=-3:5,x label style={at={(axis description cs:0.5,-0.2)},anchor=south},
            y label style={at={(axis description cs:-0.1,.5)},rotate=0,anchor=south},
            xlabel={$z$},
            ylabel={$\phi_{\!_\mathrm{ELU}}(z)$}
        ]
        \addplot+[mark=none,blue,domain=-5:0,ultra thick] {1*(exp(x)-1)};
        \addplot+[mark=none,blue,domain=0:3,ultra thick] {x};
        \addplot+[mark=none,dashed] coordinates {(-6, 0) (4, 0)};
        \addplot+[mark=none,dashed] coordinates {(0, -1) (0, 3)};
    \end{axis}
\end{tikzpicture}
\hspace{.5cm}
\centering
\begin{tikzpicture}
    \begin{axis}[scale=0.75,
        domain=-3:5,x label style={at={(axis description cs:0.5,-0.1)},anchor=north},
            y label style={at={(axis description cs:-0.1,.5)},rotate=0,anchor=south},
            xlabel={$z$},
            ylabel={$\frac{d\phi_{\!_\mathrm{ELU}}(z)}{dz}$}
        ]
        \addplot+[mark=none,blue,domain=-5:0,ultra thick] {exp(x)};
        \addplot+[mark=none,blue,domain=0:3,ultra thick] {1};
        \addplot+[mark=none,dashed] coordinates {(0, 0) (0, 1)};
        \addplot+[mark=none,dashed] coordinates {(-6, 0) (4, 0)};
    \end{axis}
\end{tikzpicture}
\caption{Exponential linear unit (ELU) and its derivative.}\label{fig:elu}
\end{figure}
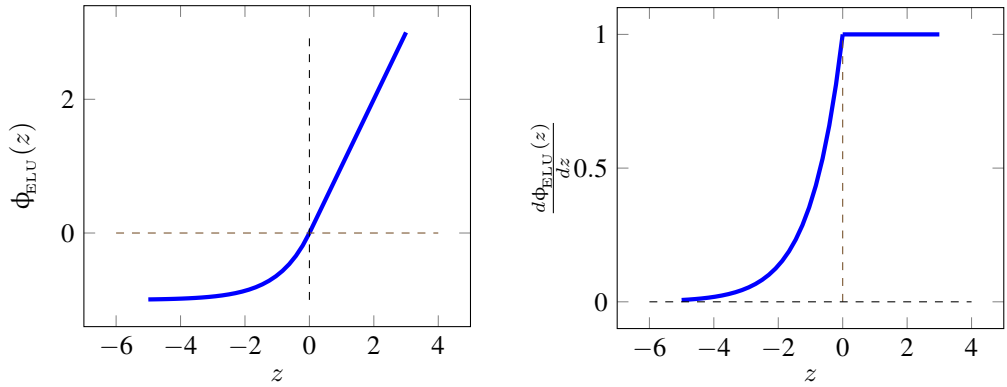

\subsubsection{Training a neural network}
\label{sec:training}
%
%


The goal of training a neural network is to estimate the vector of parameters $\ppm:= \{\Wm_i, \betaa_i\}_{i=0}^{H}$ of the network by minimizing a cost (or loss) function. Among different training procedures, \textit{e.g.}, supervised, unsupervised, or reinforcement learning \cite{goodfellow2016deep}, here we use the supervised learning with labeled training samples $\mathcal{D}:=\{(\xm_i,y_i)\}_{i=1}^{\Ntr}$ of the inputs of the QoI, where $y_i := \mathcal{M}(\xm_i)$. In this paper, we choose the mean square error (MSE) as the cost function $J$, \textit{i.e.}, 
\begin{equation}
J:=\frac{1}{\Ntr}\sum_{i=1}^{\Ntr} J_i:= \frac{1}{\Ntr}\sum_{i=1}^{\Ntr} \Big(y_i - \pinn(\xm_i;\ppm) \Big)^2.
\end{equation}
We remark that a regularization term can be added to the cost function. 

To minimize $J$, a standard stochastic gradient descent (SGD) algorithm updates the parameters at $k$th optimization step as follows
\begin{equation}
\begin{split}
\ppm_{k+1} &\leftarrow \ppm_{k} - \eta_k\frac{\partial J_{i_k}}{ \partial \ppm_{k}},\\
\end{split}
\end{equation}
where $\eta_k$ is the learning rate at $k$th iteration, $i_k$ is selected uniformly at random from $\{1,\dots,\Ntr\}$, and the derivative $\frac{\partial J_{i_k}}{ \partial \ppm_{k}}$ is calculated using back propagation \cite{goodfellow2016deep,higham2018deep}. We, however, use an improved variant of SGD, namely, the Adaptive Moment Estimation (Adam) algorithm \cite{kingma2014adam,de2019topology}. Adam leverages past gradient information to retard the descent along large gradients. This information is stored in the momentum vector $\mm$ and squared gradient vector $\vm$ as
\begin{equation}
\label{eqn:adam_updates1}
\begin{split}
\mm_{k} &= b_m \mm_{k-1} + (1-b_m) \frac{\partial J_{i_k}}{\partial \ppm_{k}}; \quad\widehat{\mm}_{k} = \frac{\mm_{k}}{1-b_m^k};\\
\vm_{k} &= b_v \vm_{k-1}+(1-b_v)\left[\frac{\partial J_{i_k}}{\partial \ppm_{k}}\right]^2; \quad \widehat{\vm}_{k}=\frac{\vm_{k}}{1-b_v^k},\\
\end{split}
\end{equation}
where $\left[\frac{\partial J_{i_k}}{\partial \ppm_{k}}\right]^2$ is performed element-wise, and $b_m$ and $b_v$ are parameters with default values 0.9 and 0.999, respectively. In (\ref{eqn:adam_updates1}), $\widehat{\mm}_{k}$ and $\widehat{\vm}_{k}$ are the unbiased momentum and  squared gradient vectors, respectively. The gradient descent step is applied next as follows
\begin{equation}
\label{eqn:adam_updates2}
\ppm_{k+1} = \ppm_{k} - \eta_k \frac{\widehat{\mm}_{k}}{\sqrt{\widehat{\vm}_{k}}+\epsilon},
\end{equation}
where the above update is performed element-wise and $\epsilon$ is a small number to avoid division by zero. We use this algorithm to train the neural networks of this study. An illustration of the steps of this algorithm is shown in Algorithm \ref{alg:adam}.

\begin{algorithm}
	\caption{\textit{Adam} \citep{kingma2014adam}}
	\label{alg:adam}
	     \textbf{Network}: $\pinn(\cdot;\ppm_0)$\;
		 \textbf{Given}: $\{\eta_k\}_{k=1}^{N_{\max}}$, $b_m$, $b_v$, and $\epsilon$\;
		 Initialize $\ppm_{1}=\ppm_0$\;
         Initialize $\mm_0 = \mathbf{0}$\;
         Initialize $\vm_0 = \mathbf{0}$\;
		\For {$k=1,2,\dots,N_{\max}$}{
		Draw $i_k$ uniformly at random from $\{1,\dots,N_{\mathrm{tr}}\}$\;
		 Compute $\frac{\partial J_{i_k}}{\partial \ppm_{k}}$\;
		 Set $\mm_k \leftarrow b_m\mm_{k-1} + (1-b_m)\frac{\partial J_{i_k}}{\partial \ppm_{k}}$\; 
		 Set $\vm_k \leftarrow b_v\vm_{k-1} + (1-b_v)\left[\frac{\partial J_{i_k}}{\partial \ppm_{k}}\right]^2\qquad$(element-wise)\;
		 Set $\widehat\mm_k \leftarrow \mm_k/(1-b_m^k)$\;
		 Set $\widehat\vm_k \leftarrow \vm_k/(1-b_v^k)$\;
		 Set $\ppm_{k+1} \leftarrow \ppm_{k} - \eta_k\frac{\widehat{\mm}_k}{\sqrt{\widehat{\vm}_k}+{\epsilon}}\qquad $(element-wise)\;} 
         \textbf{Trained network}: $\pinn(\cdot;\ppm_{\!_{N_{\max}}})$\;
\end{algorithm}

\noindent {\bf Remark.} The universal approximation theorem \cite{cybenko1989approximation,hornik1989multilayer,hornik1990universal} states that FFNs with at least one hidden layer and large enough number of neurons, and differentiable activation functions, can approximate any continuous function on a compact support. Recently, Hanin \cite{hanin2017universal} has derived expressions for the minimum number of hidden layers and neurons per layer required to represent a function within a prescribed error when ReLU activation functions are used. Despite these theoretical guarantees, the training algorithms might not be able to find the optimal values of the network parameters to achieve a desired accuracy. In practice, with small number of hidden layers, a large number of neurons per hidden layer might be needed. Additionally, often more than one layer with a small number of neurons per layer are used. In the numerical examples of this manuscript, we use a maximum of 50 neurons per layer. A new hidden layer is added to the network if the validation error is reduced with the added layer.


\subsection{Transfer Learning with Bi-fidelity Data}
\label{sec:bi_fidelity_tl}

Transfer learning for neural networks starts from a network that has already been trained with data from a similar classification or regression problem. This scenario, also known as the inductive transfer learning, is directly applicable to the training of neural networks for uncertainty propagation using bi-fidelity data. In more detail, we first train a network using the low-fidelity training dataset $\Dl=\{(\xm_{l,i},y_{l,i})\}_{i=1}^{N_l}$ to generate a surrogate model of the QoI given by the low-fidelity model. We subsequently {\it adapt} this network based on the (smaller) high-fidelity training dataset $\Dh=\{(\xm_{h,i},y_{h,i})\}_{i=1}^{N_h}$. Notice that we assume both low- and high-fidelity models have the same inputs $\xm$, and that $\xm_{l,i}$ and $\xm_{h,i}$ are samples of $\xm$ used in the low- and high-fidelity simulations. The network adaptation may be performed in multiple ways. One approach is to fix both the architecture and parameters of the low-fidelity network and train (via high-fidelity data) a network with a smaller architecture (hence fewer parameters) that maps the output of the low-fidelity network to the high-fidelity QoI. This approach parallels the bi-fidelity Gaussian process regression (or co-kriging) technique of \cite{Kennedy00,Qian06} and relies on the assumption that the map between the low- and high-fidelity QoI is {\it simpler} to learn than the one between the model inputs and the QoI. Another approach is to fix the architecture of the low-fidelity network and only update the network parameters $\ppm$ associated with the entire network or the last few hidden layers using the high-fidelity data, see, {\it e.g.}, \cite{pan2010survey,dehghani2017neural} and the references therein. As we shall explain below, both approaches are pursued in this work. 


\subsubsection{Bi-fidelity Transfer Learning with Partial Network Adaptation/Extension (BFTL)}
\label{sec:BFTL}


Here, we consider two versions of transfer learning focusing on partial adaption or extension of the low-fidelity network. In the first approach, referred to as BFTL-1, we only update the parameters of the low-fidelity network associated with the upper layers using the high-fidelity data. This modification is done by running additional (Adam) optimization steps -- in a manner similar to (\ref{eqn:adam_updates2}) -- until convergence. In our second approach, denoted as BFTL-2, a shallow network is trained to map the output of the low-fidelity network to the high-fidelity QoI. Notice the second network is kept shallow as only a smaller high-fidelity dataset is available for training. Unlike in the first approach, here the architecture and parameters of the low-fidelity network are fixed. Figure \ref{fig:tf_dnn} shows the two networks used for the second approach above, where $y_l$ and $y_h$  are the outputs of the low-fidelity network and the high-fidelity model, respectively. We note that neither BFTL-1 nor BFTL-2 does require $\{\xm_{h,i}\}_{i=1}^{N_h}\subseteq \{\xm_{l,i}\}_{i=1}^{N_l}$, thus making the transfer learning applicable to scenarios where (independent) training data from each model is available.


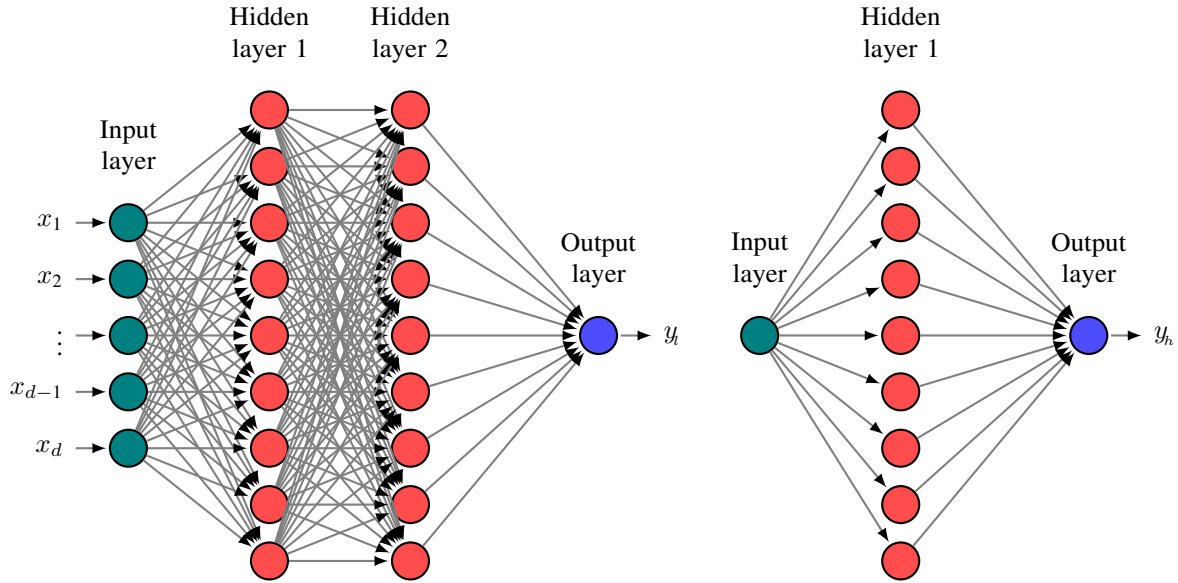
\begin{figure}
    \centering
\begin{tikzpicture}[shorten >=1pt,->,draw=black!50, node distance=\layersep,thick,latex-,scale=0.75]
    \tikzstyle{every pin edge}=[<-,shorten <=1pt,thick,latex-]
    \tikzstyle{neuron}=[circle,fill=black,minimum size=14pt,inner sep=0pt,thick,latex-]
    \tikzstyle{input neuron}=[neuron, draw=black,fill=green!50!blue,];
    \tikzstyle{output neuron}=[neuron, draw=black, fill=blue!70,thick,-latex];
    \tikzstyle{hidden neuron}=[neuron, draw=black, fill=red!70,thick,-latex];
    \tikzstyle{annot} = [text width=4em, text centered]

    \foreach \name / \y in {1,...,2}
        \node[input neuron, thick,-latex,pin=left:$x_{\y}$] (I-\name) at (0,-\y-3) {};
    \foreach \name / \y in {3,...,3}
            \node[input neuron, thick,-latex,pin=left:$\vdots$] (I-\name) at (0,-\y-3) {};
    \foreach \name / \y in {4,...,4}
            \node[input neuron, thick,-latex,pin=left:$x_{d-1}$] (I-\name) at (0,-\y-3) {};    
    \foreach \name / \y in {5,...,5}
            \node[input neuron, thick,-latex,pin=left:$x_{d}$] (I-\name) at (0,-\y-3) {};    

    \foreach \name / \y in {1,...,9}
        \path[yshift=-1cm]
            node[hidden neuron] (H1-\name) at (\layersep,-\y cm) {};

      \foreach \name / \y in {1,...,9}
        \path[yshift=-1cm]
            node[hidden neuron] (H2-\name) at (2*\layersep,-\y cm) {};

     
    \node[output neuron,-latex,pin={[pin edge={-latex}]right:${y}_{\!_l}$}, right of=H2-5] (O-1) {};
   
    \foreach \source in {1,...,5}
        \foreach \dest in {1,...,9}
            \path[thick,-latex] (I-\source) edge (H1-\dest);

  \foreach \source in {1,...,9}
        \foreach \dest in {1,...,9}
            \path[thick,-latex] (H1-\source) edge (H2-\dest);

    \foreach \source in {1,...,9}
      \foreach \dest in {1,...,1}
         \path[thick,-latex] (H2-\source) edge (O-\dest);

    \node[annot,above of=H1-1, node distance=1cm] (hl1) {Hidden layer 1};
    \node[annot,above of=H2-1, node distance=1cm] (hl2) {Hidden layer 2};
    \node[annot,above of=I-1, node distance=1cm] {Input layer};
    \node[annot,above of=O-1, node distance=1cm] {Output layer};
\end{tikzpicture}
\begin{tikzpicture}[shorten >=1pt,->,draw=black!50, node distance=\layersep,thick,latex-,yshift=-1cm,scale=0.75]
    \tikzstyle{every pin edge}=[<-,shorten <=1pt,thick,latex-]
    \tikzstyle{neuron}=[circle,fill=black,minimum size=14pt,inner sep=0pt,thick,latex-]
    \tikzstyle{input neuron}=[neuron, draw=black,fill=green!50!blue,];
    \tikzstyle{output neuron}=[neuron, draw=black, fill=blue!70,thick,-latex];
    \tikzstyle{hidden neuron}=[neuron, draw=black, fill=red!70,thick,-latex];
    \tikzstyle{annot} = [text width=4em, text centered]
    
    \foreach \name / \y in {1,...,1}
        \node[input neuron, thick,-latex,yshift = 0 cm] (I-\name) at (0,-\y+0.5) {};
    \foreach \name / \y in {1,...,9}
        \path[,yshift = 1.5 cm]
            node[hidden neuron] (H1-\name) at (\layersep,-\y+3) {};

     
    \node[output neuron,-latex,pin={[pin edge={-latex}]right:${y}_{\!_h}$}, right of=H1-5,xshift=-0*\layersep,yshift = 0 cm] (O-1) {};
   
    \foreach \source in {1,...,1}
        \foreach \dest in {1,...,9}
            \path[thick,-latex] (I-\source) edge (H1-\dest);

    \foreach \source in {1,...,9}
      \foreach \dest in {1,...,1}
         \path[thick,-latex] (H1-\source) edge (O-\dest);

    \node[annot,above of=H1-1, node distance=1cm] (hl1) {Hidden layer 1};
    \node[annot,above of=I-1, node distance=1cm] {Input layer};
    \node[annot,above of=O-1, node distance=1cm] {Output layer};
\end{tikzpicture}
    \caption{Schematic of the network in the bi-fidelity transfer learning approach BFTL-2 introduced in Section \ref{sec:BFTL}. The output of the low-fidelity network is mapped to that of the high-fidelity model via a shallow network trained using only high-fidelity data.}
    \label{fig:tf_dnn}
\end{figure}

\subsubsection{Bi-Fidelity Weighted (Transfer) Learning (BFWL)}

Following the work of \cite{dehghani2017neural,dehghani2018fidelity}, our third approach to bi-fidelity transfer learning focuses on adapting the entire low-fidelity network parameters using a synthetic dataset generated via the high-fidelity data and an adjustment of the optimization's learning rate depending on the {\it fidelity} of the synthetic data. The method is based on the concept of \textit{learning using privileged information} proposed in  \cite{vapnik2009new,vapnik2015learning}, which relies on the idea that an intelligent teacher can help educating the students. Fidelity-weighted learning (FWL), an algorithm from this paradigm, was proposed in \cite{dehghani2017neural,dehghani2018fidelity} to train neural networks for classification in the presence of training data with labels of different quality. In such instances, a neural network, known as the student network, is trained using the weakly or poorly labeled training data. A Gaussian process teacher is trained using the strongly labeled training data, which is then used to fine-tune the parameters of the student network by generating a synthetic dataset with the associated prediction confidence. See \ref{sec:gp} for more details about training a Gaussian process. A key feature of this approach is that the prediction confidence of each synthetic data entry is used to weight (or adjust) the learning rate parameter of the optimization algorithm. Since we use bi-fidelity datasets $\Dh$ and $\Dl$ within the fidelity-weighted learning technique, we name the procedure as bi-fidelity weighted learning (BFWL). The steps of BFWL are as follows:
\begin{itemize}
    \item[(1)] \textit{Stage I training}: A student neural network $\pinn ^\mathrm{s}(\xm;\ppm)$ is trained using the low-fidelity dataset $\Dl$. The network architecture can be chosen based on standard metrics, \textit{e.g.}, validation/test errors, using the relatively larger low-fidelity dataset.
    \item[(2)] \textit{Teacher construction}: The teacher, a Gaussian process $y\sim \gp (\mu, k)$ with mean function $\mu(\xm):=\mu(\widehat{y}(\xm))$ and covariance function $\kappa(\xm,\xm'):=\kappa(\widehat{y}(\xm),\widehat{y}(\xm'))$ is trained using the high-fidelity dataset $\Dh$. Here, $\widehat{y}$ denotes the $\gp$ approximation to $y$. 
    \item[(3)] \textit{Soft bi-fidelity dataset generation}: Let $\{\xm_{b,i}\}_{i=1}^{N_l+N_h}: = \{\xm_{l,i}\}_{i=1}^{N_l}\bigcup\{\xm_{h,i}\}_{i=1}^{N_h}$ be the concatenation of the low- and high-fidelity inputs. Using the $\gp$ teacher, the so-called \textit{soft bi-fidelity dataset} $\D_{b}:=\{(\xm_{b,i},\mu(\xm_{b,i}))\}_{i=1}^{N_l+N_h}$ is generated next. Associated with each $\mu(\xm_{b,i})$, the $\gp$ also provides a posterior variance $\Sigma(\xm_{b,i}) := \kappa(\xm_{b,i},\xm_{b,i})$ which will be employed in training the bi-fidelity network as explained next. 
    
%
    %
    \item[(4)] \textit{Stage II training}: To fine-tune the student network $\pinn^\mathrm{s}(\xm;\ppm)$ with help from the teacher, the parameters $\ppm$ are adapted using the Adam update (\ref{eqn:adam_updates2}) but with a weighted learning rate $\widehat{\eta}_k $,
\begin{equation}
\label{eqn:adam_bfwl}
    \ppm_{k+1} = \ppm_{k} - \widehat{\eta}_k \frac{\widehat{\mm}_{k}}{\sqrt{\widehat{\vm}_{k}}+\epsilon},
\end{equation}
where at  the $k$th iteration, $\widehat{\eta}_k$ is given by
\begin{equation}
\label{eqn:bfwl_rate}
\widehat{\eta}_k=\eta_k\exp[-\beta\Sigma(\xm_{b,i_k})].
\end{equation} 
Notice that $\widehat{\eta}_k$ accounts for the fidelity of each soft data entry via its estimated variance provided by the teacher. In (\ref{eqn:bfwl_rate}), the data index $i_k$ is selected uniformly at random from $\{1,\dots, N_l+N_h\}$ and is used to evaluate $\widehat{\mm}_{k}$ and $\widehat{\vm}_{k}$ in (\ref{eqn:adam_bfwl}); see also (\ref{eqn:adam_updates1}). Additionally, in (\ref{eqn:bfwl_rate}), $\beta$ is a positive parameter that dictates our trust in the teacher and, similarly as in (\ref{eqn:adam_updates2}), $\eta_k$ is the learning rate at iteration $k$. 
\end{itemize}




\begin{figure}
    \centering
\begin{tikzpicture}
    \node[inner sep=0pt,draw,very thick,black,rounded corners = 10,fill=black!10,minimum width = 2 cm, minimum height = 2 cm] (student) at (0,0.2)
    {\includegraphics[width=.1\textwidth]{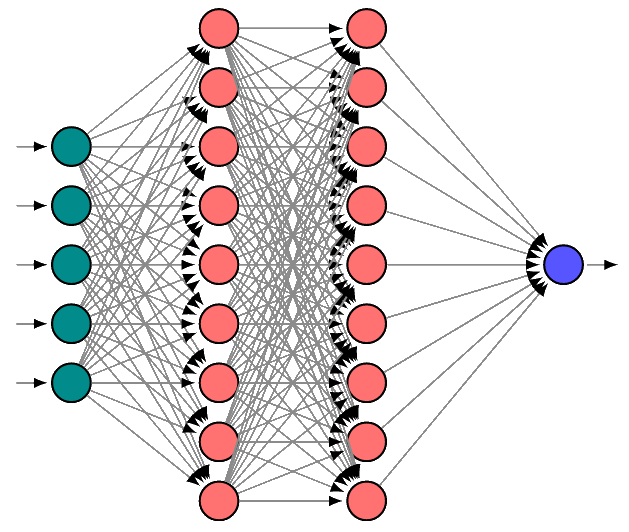}};
    \node[draw = none] at (0,1.45) {\small Student, $\pinn^\mathrm{s}$};
    \draw[-latex,thick]  (-1.7,0.2)node{} -- (-1,0.2);
    \node[draw = none,text width=2cm] at (-2.8,0.2) {\small $\Dl=\{(\xm_{l,i},y_{l,i})\}_{i=1}^{N_l}$};
    \draw[-latex,thick]  (0,-0.8) -- (0,-1.8);
    \draw[-latex,thick]  (-4.1,0.3) -- (-6.6,-2);
    \node[inner sep=0pt,draw,very thick,black!20!teal,rounded corners = 10,fill=teal!10,minimum width = 2 cm, minimum height = 2 cm] (student2) at (0,-2.8)
    {\includegraphics[width=.1\textwidth]{figs/nn_opaque_bg.png}};
    \draw[-latex,thick]  (0,-3.8) -- (0,-4.8);
    \node[inner sep=0pt,draw,very thick,black!20!green,rounded corners = 10,fill=green!10,minimum width = 2 cm, minimum height = 2 cm] (student3) at (0,-5.8)
    {\includegraphics[width=.1\textwidth]{figs/nn_opaque_bg.png}};
    \node[draw=none] at (0.7,-1.3) {\small (Adam)};
    \node[draw=none,text width = 1.5 cm] at (-0.5,-1.3) {\small Stage I training};
    \node[draw=none] at (0.7,-4.3) {\small (Adam)};
    \node[draw=none,text width = 1.5 cm] at (-0.5,-4.3) {\small Stage II training};
    
    \node[inner sep=0pt,draw,very thick,red!70,rounded corners = 10,fill=red!10,minimum width = 2 cm, minimum height = 2 cm] (student) at (-7.6,-2.8)
    {\includegraphics[width=.1\textwidth]{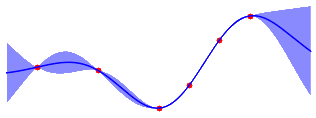}};
    \draw[-latex,thick]  (-6.6,-2.8) -- (-5.8,-2.8);
    
    \draw[-latex,thick]  (-2.2,-2.8) -- (-1,-2.8);
    \node[text width=2cm] at (-4.5,-2.8) {\small $\D_{b}:=\{(\xm_{b,i},\mu(\xm_{b,i}))\}_{i=1}^{N_l+N_h}$};
    \node[draw = none] at (-7.6,-1.55) {\small Teacher, $\gp$};
    \node[draw = none,text width = 2cm] at (-10.2,-2.8) {\small $\Dh=\{\xm_{h,i},y_{h,i}\}_{i=1}^{N_h}$};
    \draw[-latex,thick]  (-9.1,-2.8)node{} -- (-8.6,-2.8);
    \node[draw=none] at (-7.6,-3.3) {\small $\mu(\cdot),\Sigma(\cdot)$};
    
    \node[draw=none] at (2.8,-1.5) {\small Prediction error};
    

    \node[draw,fill=black!20!teal,rectangle,minimum height = 1.4 cm] at (2.6,-2.9) {~};
    \draw[-latex,very thick] (2,-3.6) -- (2,-2);
    \draw[-latex,very thick] (2,-3.6) -- (3.6,-3.6);

    \node[draw,fill=black!20!teal,rectangle,minimum height = 1.4 cm] at (2.5,-5.9) {~};
    \node[draw,fill=green!40,rectangle,minimum height = 0.4 cm] at (2.9,-6.4) {~};
    \draw[-latex,very thick] (2,-6.6) -- (2,-5);
    \draw[-latex,very thick] (2,-6.6) -- (3.6,-6.6);
    
\end{tikzpicture}
    \caption{Schematic of the fidelity weighted transfer learning, \cite{dehghani2017neural,dehghani2018fidelity}, using a bi-fidelity dataset (BFWL). The student network is trained using the low-fidelity dataset $\Dl$ (Stage I) and is then updated using the soft bi-fidelity dataset $\D_b$ (Stage II). A $\gp$ model is trained using the high-fidelity data $\Dh$ and  employed to generate $\D_b$. The posterior variance of the soft dataset is used to adjust the learning rate of the Stage II optimization solver to adapt the student network parameters using $\D_b$.}
    \label{fig:bfwl}
\end{figure}

A number of remarks regarding the BFWL algorithm above are worthwhile mentioning. First, the utility of the $\gp$ model is due to its ability to provide an estimate of the variance of the teacher's prediction of the QoI corresponding to the low-fidelity inputs $\{\xm_{l,i}\}$. This is in turn used to weight the learning rate in (\ref{eqn:bfwl_rate}).  Other types of  models, \textit{e.g.}, Bayesian variable regression \cite{mitchell1988bayesian} with polynomial basis, may be used as long as the uncertainty associated with the teacher's predictions can be quantified. Secondly, the Stage II updates (\ref{eqn:adam_bfwl}) are started from those of the learned student network in Stage I training, hence a {\it warm start} for the  Stage II training. Third, for larger values of $\beta$ in (\ref{eqn:bfwl_rate}), the parameters of the student network are less affected by the stage II updates. A detailed investigation of the effect of $\beta$, however, is beyond the scope of the current paper. Last but not least, we remark that the stage II training resemblances closely the variance reduced gradient descent algorithms, \textit{e.g.}, SVRG \cite{johnson2013accelerating} and BF-SVRG \cite{de2019bifidelity}. As an example, a control variate constructed from the low-fidelity dataset $\Dl$ for the gradients can be constructed to produce a variance reduced gradient estimate. However, the application of this variance reduction approach to the examples of Section \ref{sec:ex} did not show notable improvements in the validation error as compared to that of the standard training using only high-fidelity dataset. Therefore, the results from such an approach is included in Section \ref{sec:ex}. 
A schematic illustrating the steps of the BFWL algorithm is shown in Fig. \ref{fig:bfwl}, in which the third column shows that the performance of the student network $\pinn^\mathrm{s}$ improves after the network parameter adaption in Stage II. 



\section{Numerical Examples}\label{sec:ex}

The numerical examples presented in this section are implemented using PyTorch \cite{paszke2017automatic} for modeling neural networks and Scikit-learn \cite{pedregosa2011scikit} for Gaussian processes. In this section, we apply the bi-fidelity methods of Section \ref{sec:bi_fidelity_tl} for the training of neural networks for three scientific problems from different domains, namely, structural mechanics, electrochemistry and transport, and fluid mechanics. We compare the results to neural networks trained using only the high-fidelity dataset. 
For comparison, we use a separate high-fidelity dataset $\D^\mathrm{v}:=\{(\xm^\mathrm{v}_i,y_{h,i}^\mathrm{v})\}_{i=1}^{N_v}$ for validation and define the validation error as the normalized root mean square error (RMSE) 
\begin{equation}\label{eq:val_rmse}
     \text{RMSE} =\sqrt{\frac{\sum_{i=1}^{N_{v}} \left(y_{h,i}^\mathrm{v} - \pinn(\xm_i^\mathrm{v} )\right)^2}{\sum_{i=1}^{N_{v}} \left(y_{h,i}^\mathrm{v}\right)^2}}.
\end{equation}

%
%
\subsection{Example I: Deflection of a Composite Beam}

A cantilever beam with composite cross-section and hollow web is used for the first example as shown in Fig. \ref{fig:ex1}. The dimensions of the beam are as follows:  the length $L=50$\,m; the width $w=1$\,m; the radius of the circular holes $r=1.5$\,m; and the heights of different parts of the web $h_1=h_2=0.1$\,m, and $h_3 = 5$\,m. The externally distributed load $q$ and the elastic moduli $E_1$, $E_2$, and $E_3$ of the three materials constituting the beam cross-section are assumed to be independent random variables uniformly distributed, as specified in Table \ref{tab:beam_unc}. Thus, the input $\xm=(q,E_{1},E_{2},E_{3})$. The QoI $y$ is the maximum vertical displacement (at the free end of the beam). A finite element model for the beam is implemented using FEniCS \cite{LoggMardalEtAl2012a} and used as the high-fidelity model. The mesh for this high-fidelity model is shown in Fig. \ref{fig:ex1_hf}. For the low-fidelity model, the deflection of the beam $u_l(x)$ is given by the Euler-Bernoulli beam theory while ignoring the circular holes,
\begin{equation}\label{eq:beam}
\begin{split}
    & EI\frac{\mathrm{d}^4u_l(x)}{\mathrm{d}x^4} = -q,\\
    & \text{subject to }u_l(0) = 0;~ \frac{du_l(0)}{dx} = 0; ~ \frac{d^3u_l(L)}{dx^3}=0;~\frac{d^4u_l(L)}{dx^4}=0,\\
\end{split}
\end{equation}
where $E$ and $I$ are the Young's modulus and the moment of inertia of an {\it equivalent} cross-section consisting of a single (homogenized) material. The solution of (\ref{eq:beam}) can be written as
\begin{equation}\label{eq:beam_lf}
    u_l(x) = -\frac{qL^4}{24EI}\left[ \left(\frac{x}{L}\right)^4 - 4\left(\frac{x}{L}\right)^3 + 6\left(\frac{x}{L}\right)^2 \right].
\end{equation}
Note that for the low-fidelity model, the effect of the uncertainty is only through a multiplicative factor in (\ref{eq:beam_lf}). However, this is not the case for the high-fidelity model. 

\begin{figure}[!htb]
    \centering
    \includegraphics[scale=1.5]{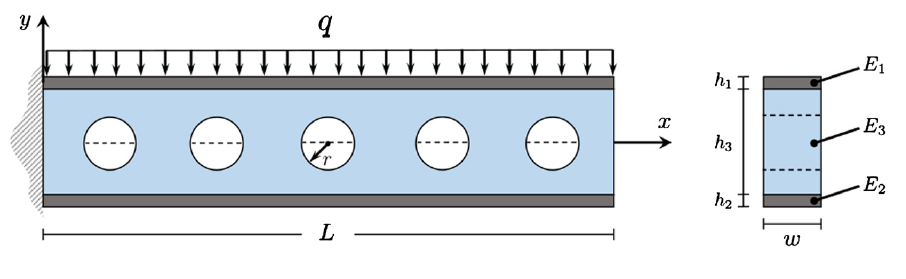}
    \caption{The composite beam used in Example I (adapted from \cite{hampton2018practical}). The dimensions are as follows: $L=50$\,m, $r=1.5$\,m, $h_1=h_2=0.1$\,m, and $h_3=5$\,m. The elastic moduli $E_1$, $E_2$, $E_3$, and the external load $q$ are assumed uncertain.}
    \label{fig:ex1}
\end{figure}

\begin{table}[!htb]
 \caption{The uncertain parameters in Example I and their corresponding probability distributions. Note that $\mathcal{U}[a,b]$ denotes a uniform distribution between $a$ and $b$.}
    \label{tab:beam_unc}
    \centering
    \begin{tabular}{l|c|c}
    \hline \rule{0pt}{2ex} 
       Parameter  & Unit & Distribution \\
       \hline
\rule{0pt}{2ex}         Distributed load, $q$ & kN/m & $\mathcal{U}[9,11]$\\
        Elastic modulus $E_1$ & MPa & $\mathcal{U}[0.9, 1.1]$\\
        Elastic modulus $E_2$ & MPa & $\mathcal{U}[0.9, 1.1]$\\
        Elastic modulus $E_3$ & kPa & $\mathcal{U}[9, 11]$ \\\hline
    \end{tabular}
\end{table}

\begin{figure}[!htb]
    \centering
    \includegraphics[scale=1.5]{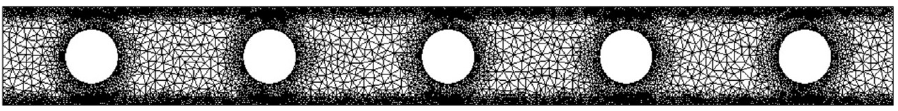}
    \caption{The composite beam is solved in FEniCS and used as the high-fidelity model with the mesh shown here.}
    \label{fig:ex1_hf}
\end{figure}

\subsubsection{Results}

We use a neural network that consists of two hidden layers with 15 neurons each and ELU activation function with $\alpha=1$ based on a preliminary investigation, where we monitored the validation error as we increased the number of neurons per layer (see Fig. \ref{fig:beam_neurons}). Note that we similarly choose our networks for the other two examples in this paper. 
\begin{figure}[!htb]
    \centering
    \includegraphics[scale = 0.3]{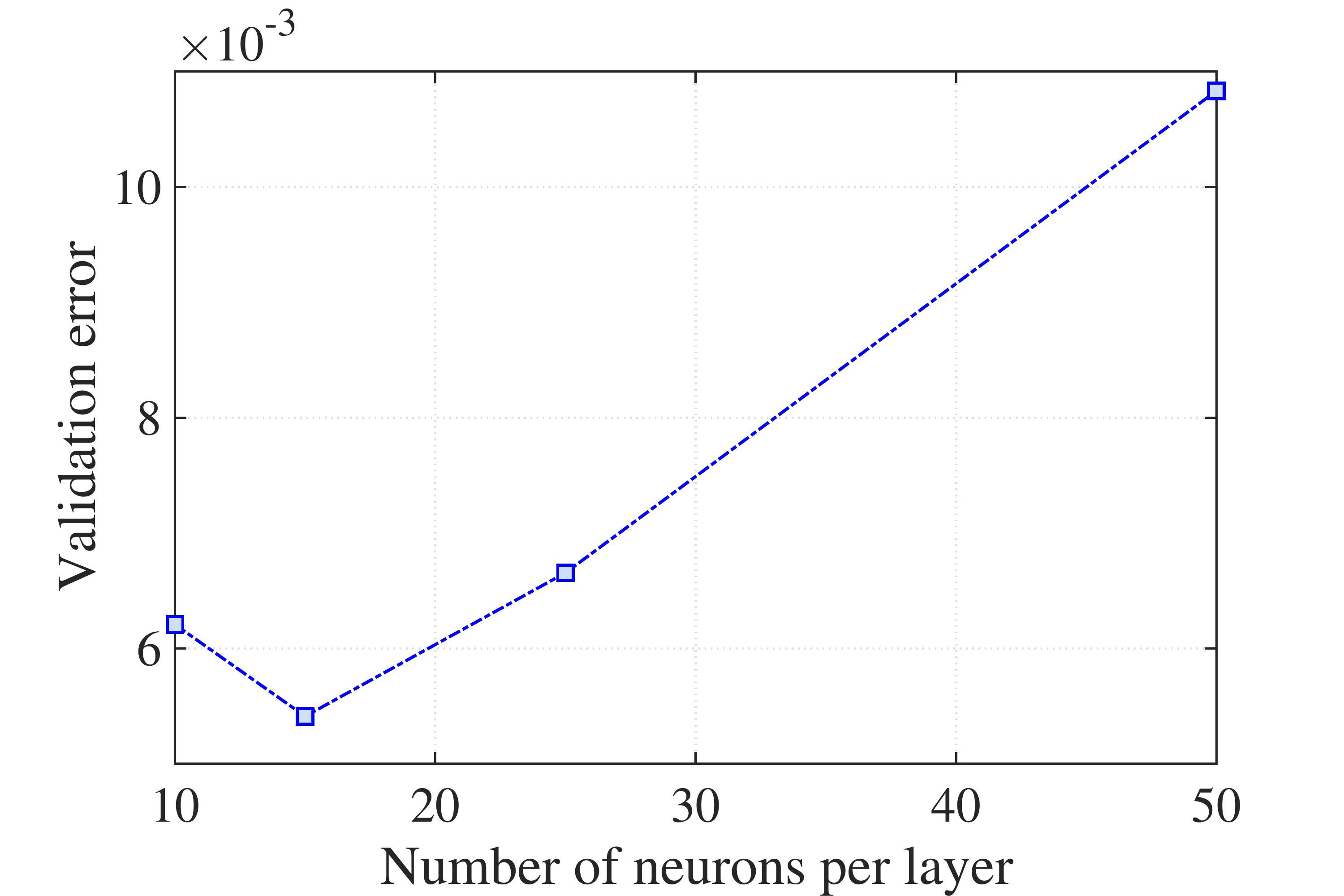}
    \caption{For a feed-forward neural network with two hidden layers the number of neurons (kept same for these two layers) is selected based on the smallest validation RMSE. The validation errors are displaced here for Example I.}
    \label{fig:beam_neurons}
\end{figure}
We train the FNN and ResNet networks with $N_h=30$ high-fidelity samples. For validation, we use $N_v = 50$ samples and the error defined in \eqref{eq:val_rmse} to assess the network performance. 
Next, we use different transfer learning techniques with $N_l=250$ and $N_h=20$, respectively, low- and high-fidelity samples randomly generated based on the distribution of the inputs. Note that the cost of generating the training data from the low-fidelity model is negligible compared to the high-fidelity model in this example. For the BFTL-2 approach, we add a third hidden layer consisting of 20 neurons to model the relation between $y_l$ and $y_h$. For the BFWL method, we use an additive combination of radial basis kernel with length scale of 1 and white noise kernel with small noise level bounds $(10^{-3},10^{-5})$ to train the Gaussian process teacher. 
The learning rates used with different methods are listed in Table \ref{tab:beam_lr}.

\begin{table}[!htb]
 \caption{The learning rates used with different learning methods in Example I.}
    \label{tab:beam_lr}
    \centering
    \begin{tabular}{l|c|c|c}
    \hline \rule{0pt}{2ex} 
       Architecture & Method & Dataset & Learning rate, $\eta_k$\\
       \hline
\rule{0pt}{2ex}
\multirow{ 7}{*}{FNN} & Standard & $\Dh$ & $10^{-4}$\\\cline{2-4}
\rule{0pt}{2ex} & \multirow{2}{*}{BFTL-1} & $\Dl$ & $4\times10^{-4}$\\
& & $\Dh$ & $10^{-4}$\\\cline{2-4} \rule{0pt}{2ex}
& \multirow{2}{*}{BFTL-2} & $\Dl$ & $10^{-3}$\\
& & $\Dh$ & $10^{-4}$\\\cline{2-4} \rule{0pt}{2ex}
& \multirow{2}{*}{BFWL ($\pinn^\mathrm{s}$)} & $\Dl$ & $10^{-3}$\\
& & $\Dh$ & $2\times10^{-4}\quad (\beta = -0.25)$\\ \hline
\rule{0pt}{2ex}
\multirow{ 7}{*}{ResNet} & Standard & $\Dh$ & $10^{-1}$\\\cline{2-4}
\rule{0pt}{2ex} & \multirow{2}{*}{BFTL-1} & $\Dl$ & $4\times10^{-4}$\\
& & $\Dh$ & $10^{-4}$\\\cline{2-4} \rule{0pt}{2ex}
& \multirow{2}{*}{BFTL-2} & $\Dl$ & $10^{-3}$\\
& & $\Dh$ & $10^{-4}$\\\cline{2-4} \rule{0pt}{2ex}
& \multirow{2}{*}{BFWL ($\pinn^\mathrm{s}$)} & $\Dl$ & $10^{-3}$\\
& & $\Dh$ & $2\times10^{-4}\quad (\beta = -0.25)$\\
\hline
    \end{tabular}
\end{table}

\begin{figure}[!htb]
    \centering
    \includegraphics[scale = 0.55]{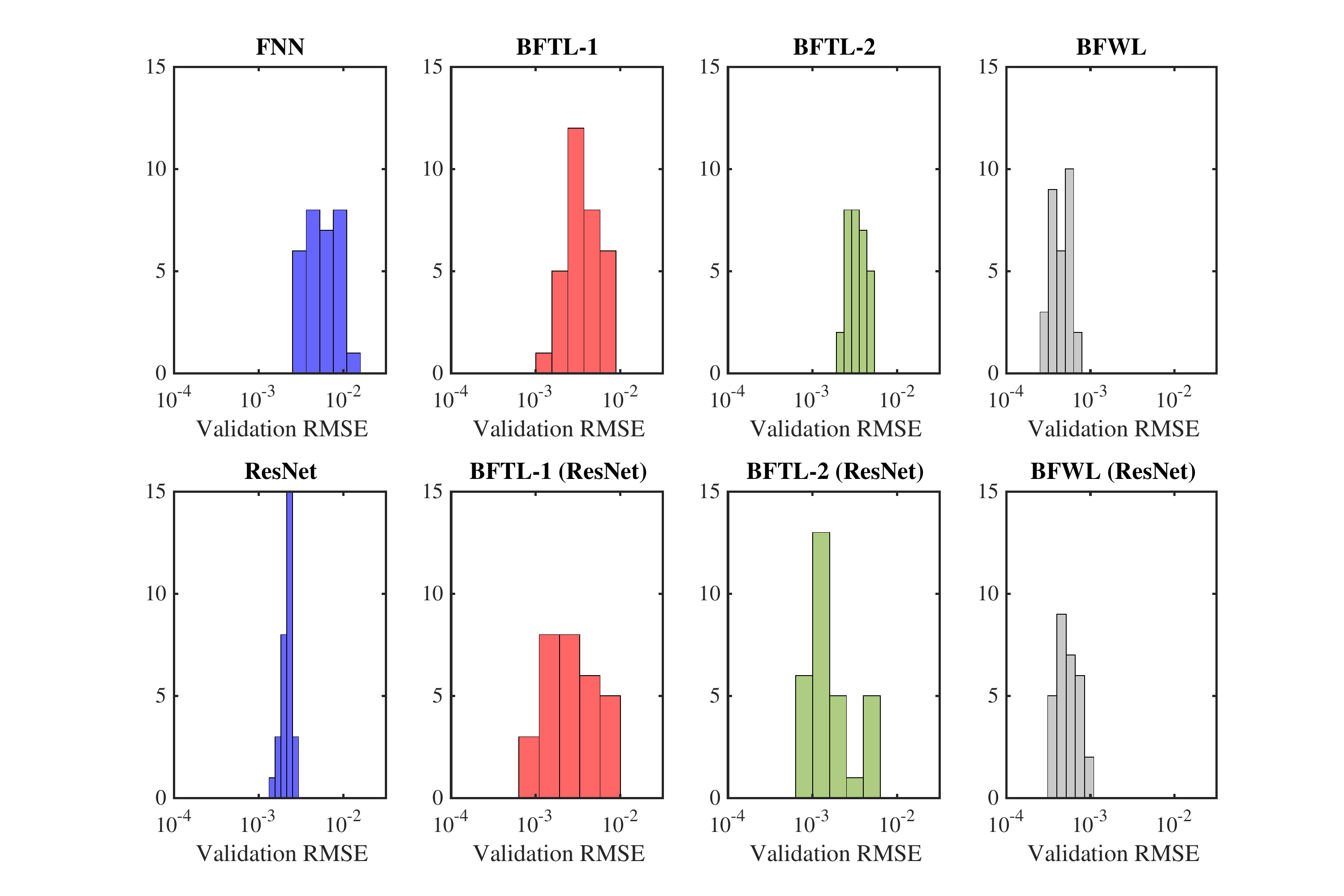}
    \caption{Histograms of the validation error RMSE for Example I using different learning approaches to train the neural network for 30 random initialization of the network. The first column shows the validation RMSEs using standard learning techniques for two different architectures using $N_h=20$ high-fidelity samples. The rest of the columns show the validation RMSEs for different transfer learning strategies implemented herein using $N_h=20$ high-fidelity and $N_l=250$ low-fidelity samples.}
    \label{fig:beam_hist1}
\end{figure}

\begin{figure}[!htb]
    \centering
    \includegraphics[scale = 0.55]{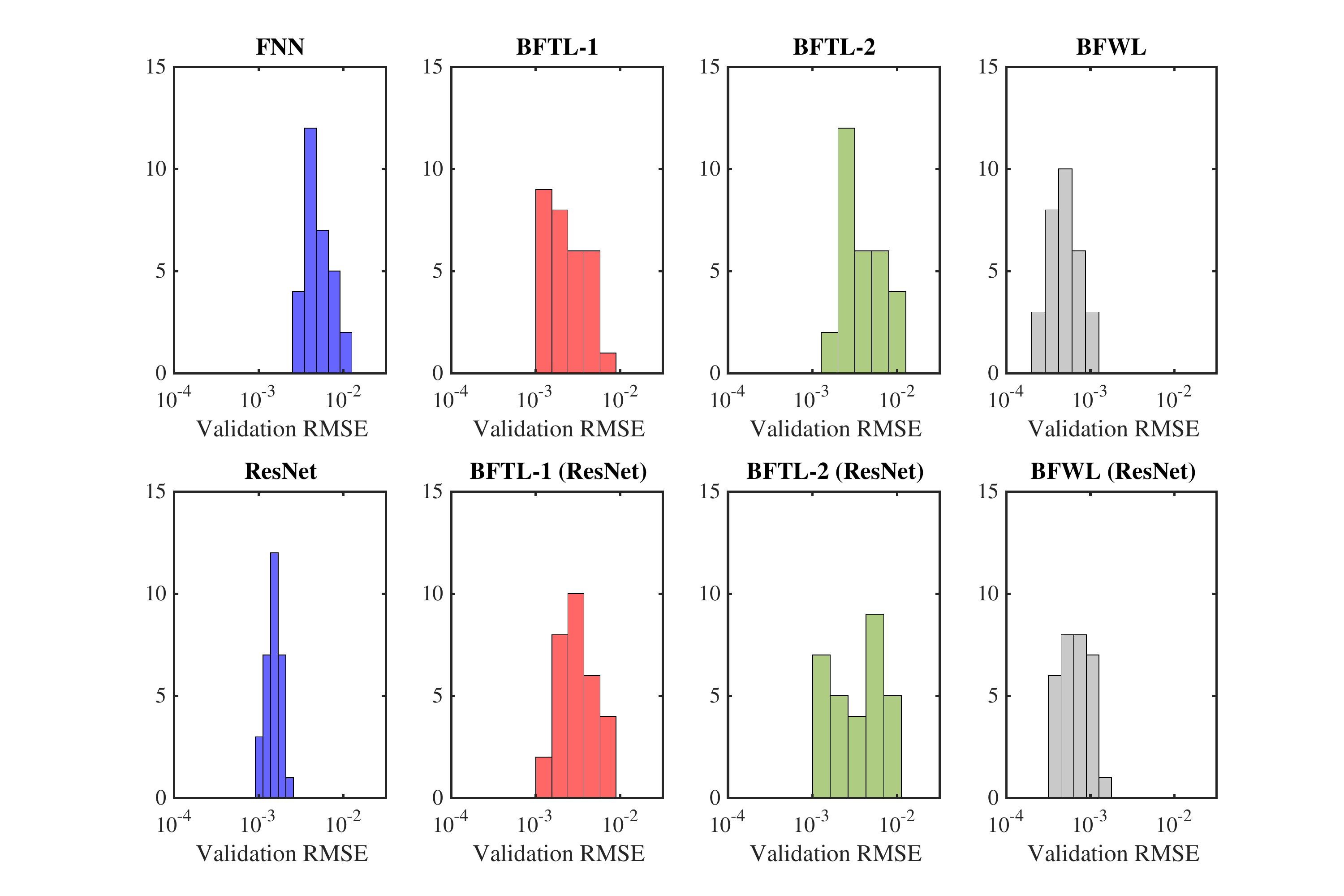}
    \caption{Histograms of the log of validation RMSEs for Example I using different learning approaches to train the neural network for 30 different validation datasets. The first column shows the validation RMSEs using standard learning techniques for two different architectures using $N_h=20$ high-fidelity sample. The rest of the columns show the validation RMSEs for different transfer learning strategies implemented herein using $N_h=20$ high-fidelity and $N_l=250$ low-fidelity samples. }
    \label{fig:beam_hist2}
\end{figure}
Tables \ref{tab:beam_mean_RMSE_I} and \ref{tab:beam_mean_RMSE_II} show the mean validation RMSEs obtained from different learning methods. To illustrate the variability of the trained network with respect to the choice of the initial parameters $\ppm$, Fig. \ref{fig:beam_hist1} shows the histograms of the validation RMSEs for 30 independent, random initializations of the network parameters but with fixed training and validation datasets. Note that if the training algorithms were able to reach the same (global) optimum, then the histograms in Fig. \ref{fig:beam_hist1} would show single bars. However, this is not the case as the cost function here is non-linear and non-convex, and may feature multiple local minima. Fig. \ref{fig:beam_hist2} displays the same quantity for a fixed choice of network initial parameters but with 30 random selection of the training ($N_l=250$ and $N_h=20$) and validation ($N_v = 50$) datasets from the same pool of samples. 
  
\begin{table}[!htb]
 \caption{The mean validation RMSEs obtained using different learning methods and 30 different initializations of the network in Example I.}
    \label{tab:beam_mean_RMSE_I}
    \centering
    \begin{tabular}{l|c|c}
    \hline \rule{0pt}{2ex} 
       \multirow{2}{*}{Architecture} & \multirow{2}{*}{Method} & Mean validation\\
       & & RMSE\\
       \hline
\rule{0pt}{2ex}
\multirow{ 4}{*}{FNN} & Standard &  $6.0055\times10^{-3}$ \\
\rule{0pt}{2ex} & {BFTL-1} & $4.0435\times10^{-3}$ \\
 \rule{0pt}{2ex}
& {BFTL-2}  & $3.4245\times10^{-3}$\\
 \rule{0pt}{2ex}
& {BFWL} & $4.5453\times10^{-4}$\\
\hline
\rule{0pt}{2ex}
\multirow{ 4}{*}{ResNet} & Standard &  $2.1694\times10^{-3}$\\
\rule{0pt}{2ex} & {BFTL-1} & $3.3929\times10^{-3}$ \\
 \rule{0pt}{2ex}
& {BFTL-2} & $1.9507\times10^{-3}$ \\
 \rule{0pt}{2ex}
& {BFWL} & $5.6579\times10^{-4}$ \\
\hline
    \end{tabular}
\end{table}
\begin{table}[!htb]
 \caption{The mean validation RMSEs obtained using different learning methods and 30 random replications of training/validation datasets in Example I.}
    \label{tab:beam_mean_RMSE_II}
    \centering
    \begin{tabular}{l|c|c}
    \hline \rule{0pt}{2ex} 
       \multirow{2}{*}{Architecture} & \multirow{2}{*}{Method} & Mean validation\\
       & & RMSE\\
       \hline
\rule{0pt}{2ex}
\multirow{ 4}{*}{FNN} & Standard & $5.4115\times10^{-3}$ \\
\rule{0pt}{2ex} & {BFTL-1} & $2.6621\times10^{-3}$ \\
 \rule{0pt}{2ex}
& {BFTL-2} & $4.5706\times10^{-3}$ \\
 \rule{0pt}{2ex}
& {BFWL} & $4.9860\times10^{-4}$ \\
\hline
\rule{0pt}{2ex}
\multirow{ 4}{*}{ResNet} & Standard & $1.5013\times10^{-3}$ \\
\rule{0pt}{2ex} & {BFTL-1} & $3.3807\times10^{-3}$ \\
 \rule{0pt}{2ex}
& {BFTL-2} & $4.2873\times10^{-3}$ \\
 \rule{0pt}{2ex}
& {BFWL} & $7.3800\times10^{-4}$ \\
\hline
    \end{tabular}
\end{table}
From Figs. \ref{fig:beam_hist1} and \ref{fig:beam_hist2}, it can be noticed that ResNet provides smaller validation errors. In particular, the mean validation RMSE reduces from $6.01\times10^{-3}$ to $2.17\times10^{-3}$ using the ResNet architecture (see Table \ref{tab:beam_mean_RMSE_I}). Similarly, as reported in Table \ref{tab:beam_mean_RMSE_II}, the mean validation RMSE with fixed initial parameters reduces from $5.41\times 10^{-3}$ to $1.50\times10^{-3}$.
Second and third columns of Fig. \ref{fig:beam_hist1} and \ref{fig:beam_hist2} show that BFTL-1 and BFTL-2 improve the performance of the neural network compared to an FNN but they do not provide significant advantage over the ResNet architecture with standard training (first column second row of Figs \ref{fig:beam_hist1} and \ref{fig:beam_hist2}). 
For example, according to Table \ref{tab:beam_mean_RMSE_I}, the mean validation RMSEs of BFTL-1 and BFTL-2 using FNN are $4.04\times10^{-3}$ and $3.43\times10^{-3}$, respectively, and using ResNet are $3.39\times10^{-3}$ and $1.95\times10^{-3}$, respectively. 
Similarly, in Table \ref{tab:beam_mean_RMSE_II}, the mean validation RMSEs of BFTL-1 and BFTL-2 using FNN are $2.66\times10^{-3}$ and $4.57\times10^{-3}$, respectively, and using ResNet are $3.38\times10^{-3}$ and $4.29\times10^{-3}$, respectively.
This shows that the ResNet architecture alone improves the performance in this example by modeling the residual in the second hidden layer. However, when used in conjunction with transfer learning techniques, modeling of residuals for the low-fidelity data does not play any role in BFTL-1. Similarly, in BFTL-2, we model a relation between the low- and high-fidelity data using an extra shallow layer. As a result, more accurate modeling of the residual for the low-fidelity dataset does not remain relevant anymore. 
The fourth columns of Figs. \ref{fig:beam_hist1} and \ref{fig:beam_hist2} show that the BFWL method with both FNN and ResNet architectures performs best among all the transfer learning methods and is able to considerably reduce the error as compared to the standard training. In Fig. \ref{fig:beam_hist1}, the BFWL with FNN and ResNet gives mean validation RMSEs $4.55\times10^{-4}$ and $5.66\times10^{-4}$, respectively (see Table \ref{tab:beam_mean_RMSE_I}). In Fig. \ref{fig:beam_hist2}, the mean validation RMSEs are $4.99\times10^{-4}$ and $7.38\times10^{-4}$, respectively (see Table \ref{tab:beam_mean_RMSE_II}). Hence, in this example, the use of uncertainty information learned from a handful of the high-fidelity data using a Gaussian process teacher provides useful information to further improve the neural network's performance compared to other transfer learning techniques. ResNet architecture does not provide notable advantage within the BFWL approach.

\begin{figure}[!htb]
    \centering
    \includegraphics[scale = 0.3]{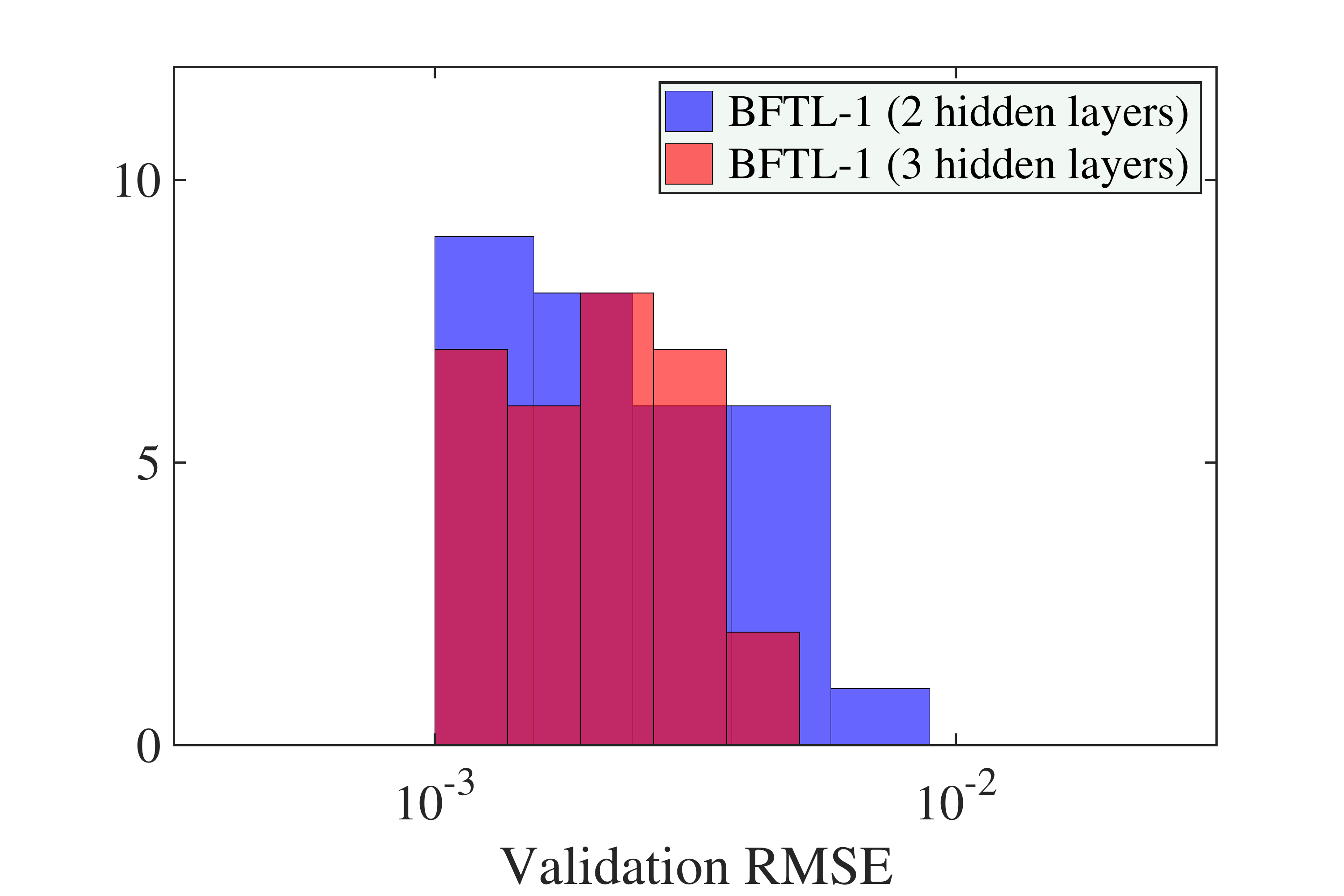}
    \caption{Comparison of histograms of validation RMSEs for BFTL-1 with two and three hidden layers (15 neurons in each) for Example I. }
    \label{fig:beam_TL1_comp}
\end{figure}

\begin{figure}[!htb]
    \centering
    \includegraphics[scale = 0.3]{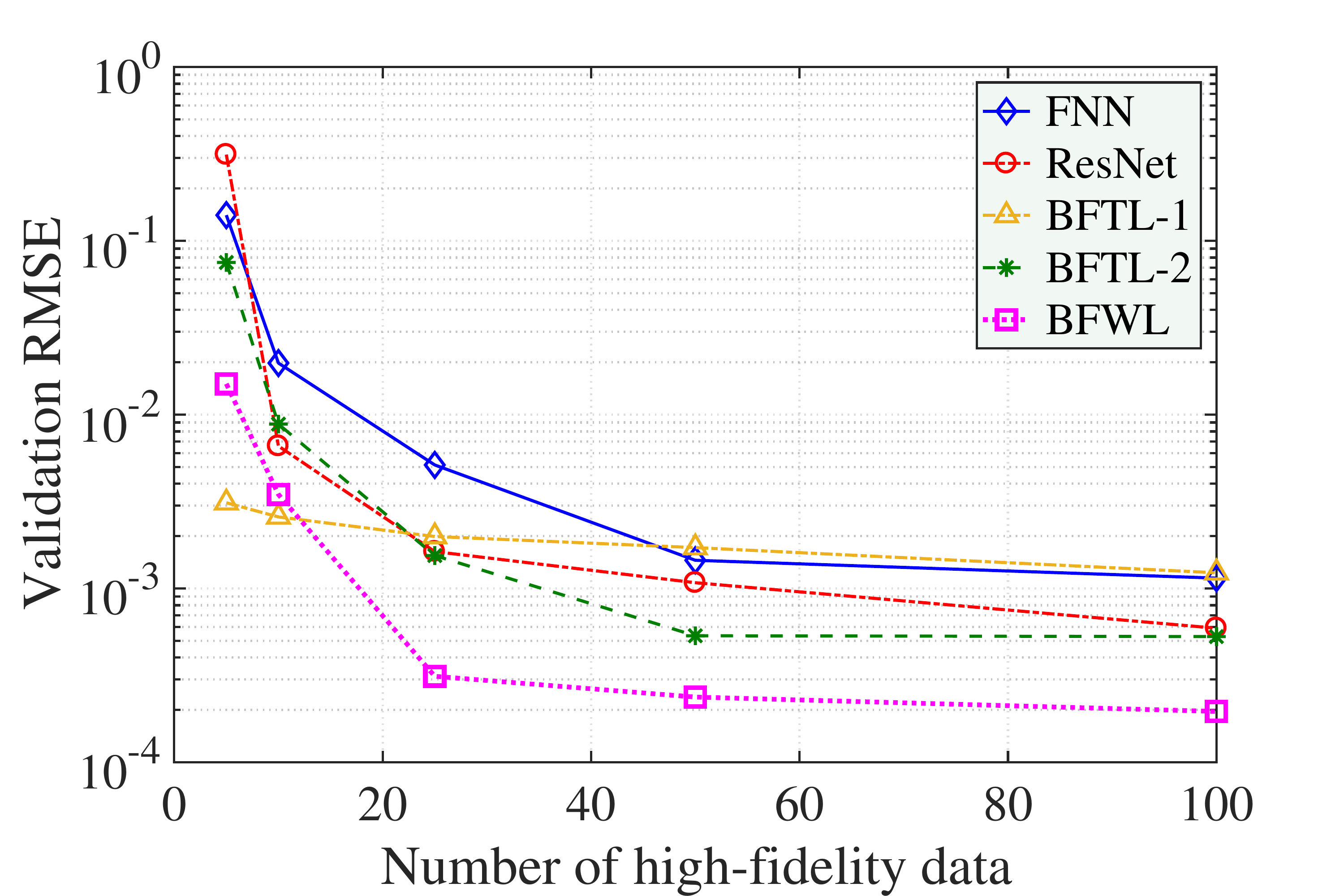}
    \caption{As more high-fidelity data are used the validation RMSE using different learning techniques reduces in Example I. }
    \label{fig:beam_Nh}
\end{figure}

\begin{figure}[!htb]
    \centering
    \includegraphics[scale = 0.3]{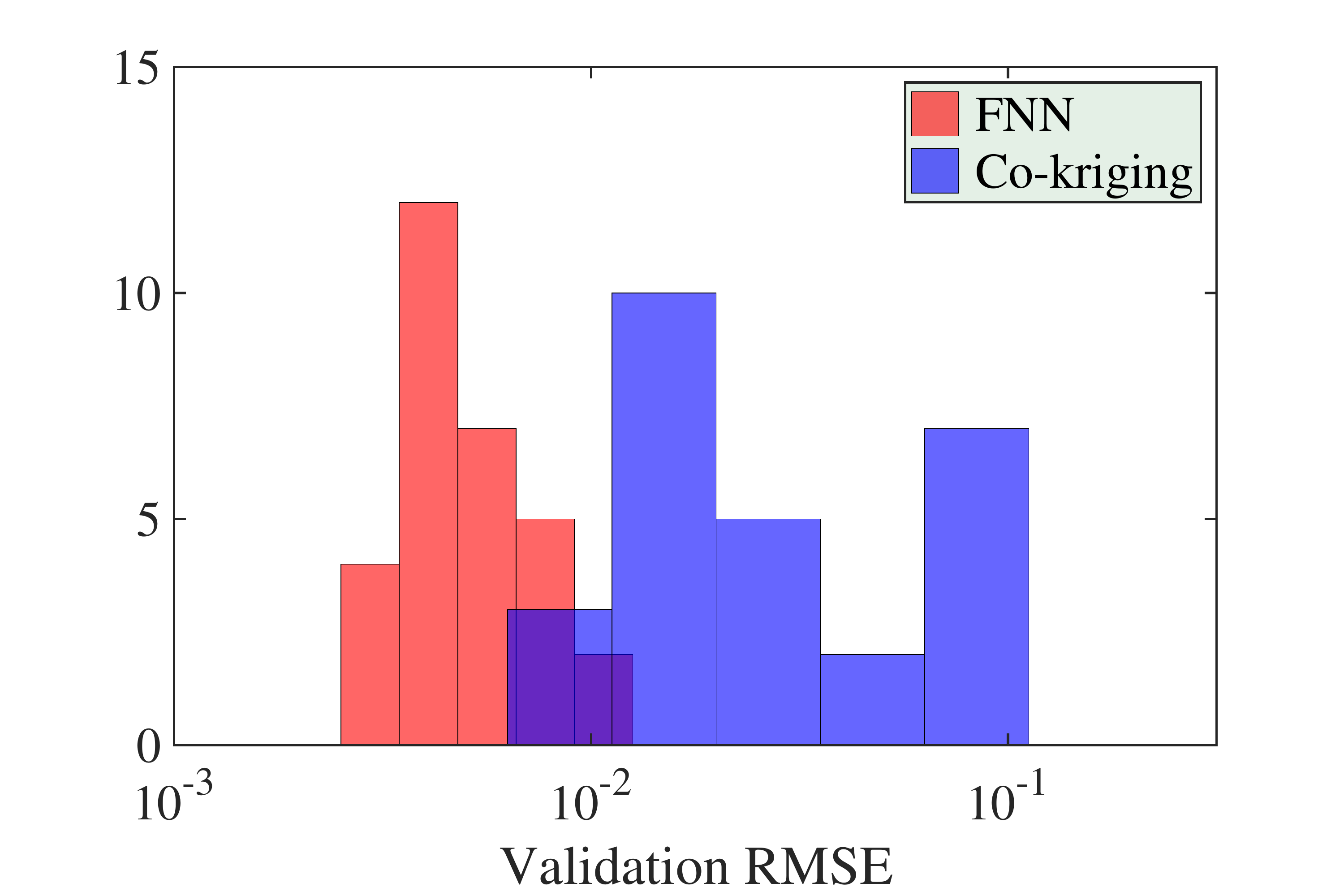}
    \caption{Comparison of co-kriging (see \ref{sec:cokriging}) and a feed-forward neural network for Example I.}
    \label{fig:beam_cokriging}
\end{figure}
We also evaluate the BFTL-1 approach with three hidden layers (15 neurons each) in this example. During the training with $\Dh$, we only train the parameters of the third hidden layer. This is to some extent similar to the BFTL-2 scenario, where a third layer is added to model the relation between $y_l$ and $y_h$. However, the BFTL-1 with three hidden layers does not show any drastic improvement in the validation RMSE; see Fig. \ref{fig:beam_TL1_comp}. Fig. \ref{fig:beam_Nh} shows a typical scenario, where we use $N_l=250$ and increase $N_h$ gradually. As the number of high-fidelity data points is increased the validation RMSE is decreased gradually similar to Fig. \ref{fig:bftl_concept}. As in the BFWL approach we rely on a Gaussian process model, a natural question is how BFWL (or FNN) compared with co-kriging (a bi-fidelity Gaussian process model).  Fig. \ref{fig:beam_cokriging} shows a comparison between FNN and co-kriging for 30 different training/validation datasets to show that in this example the FNN outperforms the co-kriging method. The description of the co-kriging approach used here is presented in \ref{sec:cokriging}.


\subsection{Example II: Lithium-ion Battery}\label{sec:exII}

\begin{figure}[!htb]
    \centering
    \includegraphics[scale=1.3]{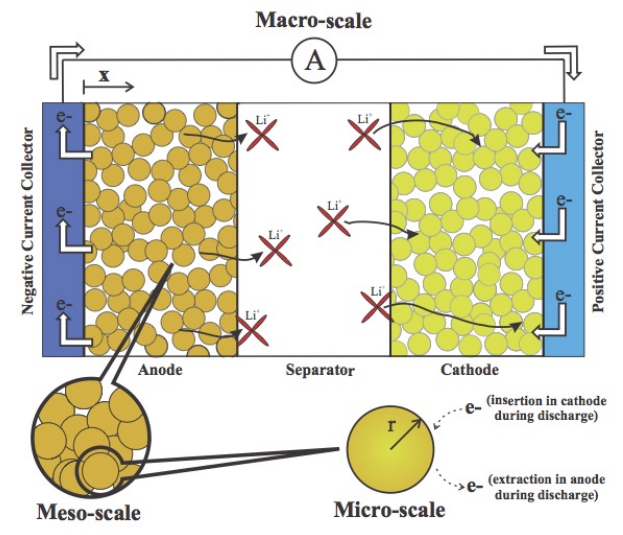}
    \caption{The schematic of a Lithium-ion battery used in Example II (adapted from \cite{hadigol2015uncertainty}).}
    \label{fig:ex2}
\end{figure}

We consider a Lithium-ion battery (LIB) model for our second example. An LIB is made up of an anode, a cathode, a separator, an electrolyte, and positive and negative current collectors; see Fig. \ref{fig:ex2}. The separator acts as a barrier between the positive electrode (anode) and the negative electrode (cathode). During the discharge of the battery, lithium ions in the anode particles diffuse to the particle surface and oxidize into positively charged lithium ions Li$^+$ and electrons. The Li$^+$ ions travel from the anode to the cathode via the electrolyte (liquid phase) using diffusion and migration, and reduce to lithium at the surface of the cathode particles before diffusing into them. During charging, the opposite happens. For the numerical example of this section, we employ the LIB model of Hadigol \textit{et al}. \cite{hadigol2015uncertainty}, which relies on the porous electrode theory of Newman in  \cite{newman1975porous}. The interested reader is referred to \cite{hadigol2015uncertainty} for a detailed description of the governing equations, model parameters, and solution strategy. In this example, we use the spatial mean of the liquid phase concentration at time $t = 2000$\,s as the QoI. A finite difference mesh with $314$ nodes is used to generate the high-fidelity dataset and a coarse mesh with $50$ nodes for the low-fidelity dataset. The coarse model runs 7 times faster than the fine one. We chose 17 parameters from \cite{hadigol2015uncertainty} to use as the input uncertain parameters $\xm$. Table \ref{tab:battery_param} shows the probability distributions of these parameters inferred from the LiC$_6$/LiCoO$_2$ cell literature; see \cite{hadigol2015uncertainty} for more details. For the interest of completeness, we also describe these parameters briefly in \ref{sec:exIIparam}. \\

\begin{table}[!htb]
\caption{The uncertain parameters used in Example II and their distributions from \cite{hadigol2015uncertainty}. A brief description of these parameters is given in \ref{sec:exIIparam}. Note that $\mathcal{U}[a,b]$ denotes a uniform distribution between $a$ and $b$.} \label{tab:battery_param}
\begin{center}
\begin{tabular}{ c|l|c|c } 
\hline\rule{0pt}{2ex} 
Component & Parameter & Unit & Distribution \rule{0pt}{2ex} \\
\hline \rule{0pt}{2ex} 
\multirow{6}{*}{Anode} & Porosity, $\epsilon_\mathrm{a}$& -- & $\mathcal{U}[0.46, 0.51]$ \\ 
& Bruggeman coeff., $b_\mathrm{a}$ & -- & $\mathcal{U}[3.8, 4.2]$ \\
& Solid diffusion coeff., $D_\mathrm{s,a}$ & m$^2$/s & $\mathcal{U}[3.51, 4.29] \times 10^{-14}$ \\
& Conductivity, $\sigma_\mathrm{a}$ & S/m & $\mathcal{U}[90, 110]$\\
& Reaction rate, $k_\mathrm{a}$ & m$^4\cdot$mol$\cdot$s & $\mathcal{U}[4.52, 5.53] \times 10^{-11}$\\
& Length, $L_\mathrm{a}$ & $\mu$m & $\mathcal{U}[77, 83]$\\
\hline \rule{0pt}{2ex} 
\multirow{6}{*}{Cathode} & Porosity, $\epsilon_\mathrm{c}$ & -- & $\mathcal{U}[0.36, 0.41]$ \\ 
& Bruggeman coeff., $b_\mathrm{c}$ & -- & $\mathcal{U}[3.8, 4.2]$\\
& Solid diffusion coeff., $D_\mathrm{s,c}$ & m$^2$/s & $\mathcal{U}[0.90, 1.10] \times 10^{-14}$ \\
& Conductivity, $\sigma_\mathrm{c}$ & S/m & $\mathcal{U}[90, 110]$\\
& Reaction rate, $k_\mathrm{c}$ & m$^4\cdot$mol$\cdot$s & $\mathcal{U}[2.10, 2.56] \times 10^{-11}$\\
& Length, $L_\mathrm{c}$ & $\mu$m & $\mathcal{U}[85, 91]$\\
\hline \rule{0pt}{2ex} 
\multirow{3}{*}{Separator} & Porosity, $\epsilon_\mathrm{s}$ & -- & $\mathcal{U}[0.63, 0.81]$\\
& Bruggeman coeff., $b_\mathrm{a}$ & -- & $\mathcal{U}[3.2, 4.8]$\\
& Length, $L_\mathrm{a}$ & $\mu$m & $\mathcal{U}[22, 28]$\\
\hline
& Li$^+$ transference number, $t_+^0$ & -- & $\mathcal{U}[0.345, 0.381]$\\
& Salt diffusion coefficient, $D$ & m$^2$/s & $\mathcal{U}[6.75, 8.25] \times 10^{-10}$\\
\hline
\end{tabular}
\end{center}
\end{table}

\begin{figure}[!htb]
    \centering
    \includegraphics[scale = 0.55]{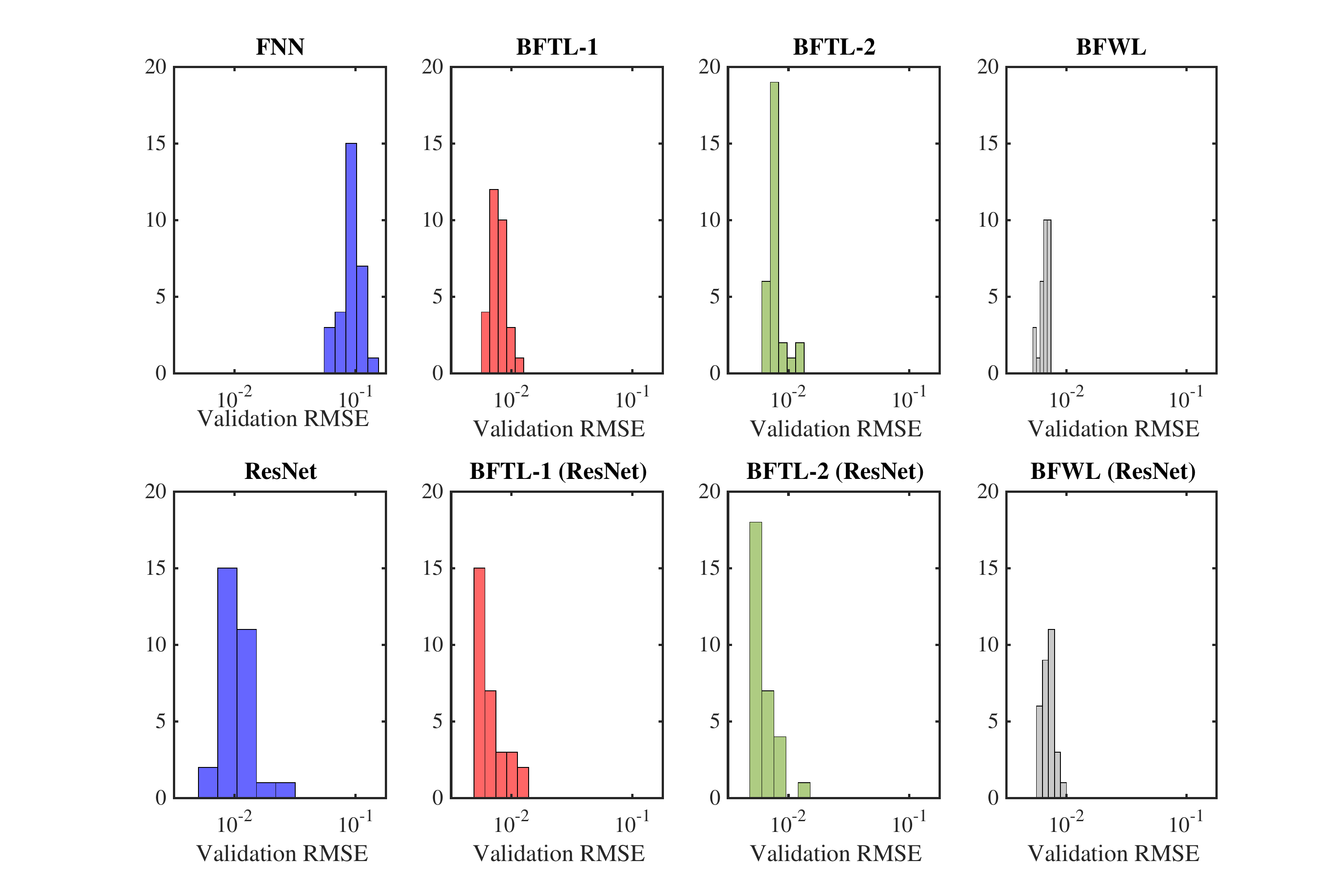}
    \caption{Histograms of validation RMSEs for Example II using different learning approaches for 30 random initializations of the network. The first column shows the validation RMSEs using standard learning technique for two different architectures and $N_h=40$ high-fidelity samples. The rest of the columns show the validation RMSEs for different transfer learning strategies implemented herein using $N_h=20$ high-fidelity and $N_l=140$ low-fidelity samples.
    }
    \label{fig:battery_hist1}
\end{figure}

\begin{figure}[!htb]
    \centering
    \includegraphics[scale = 0.55]{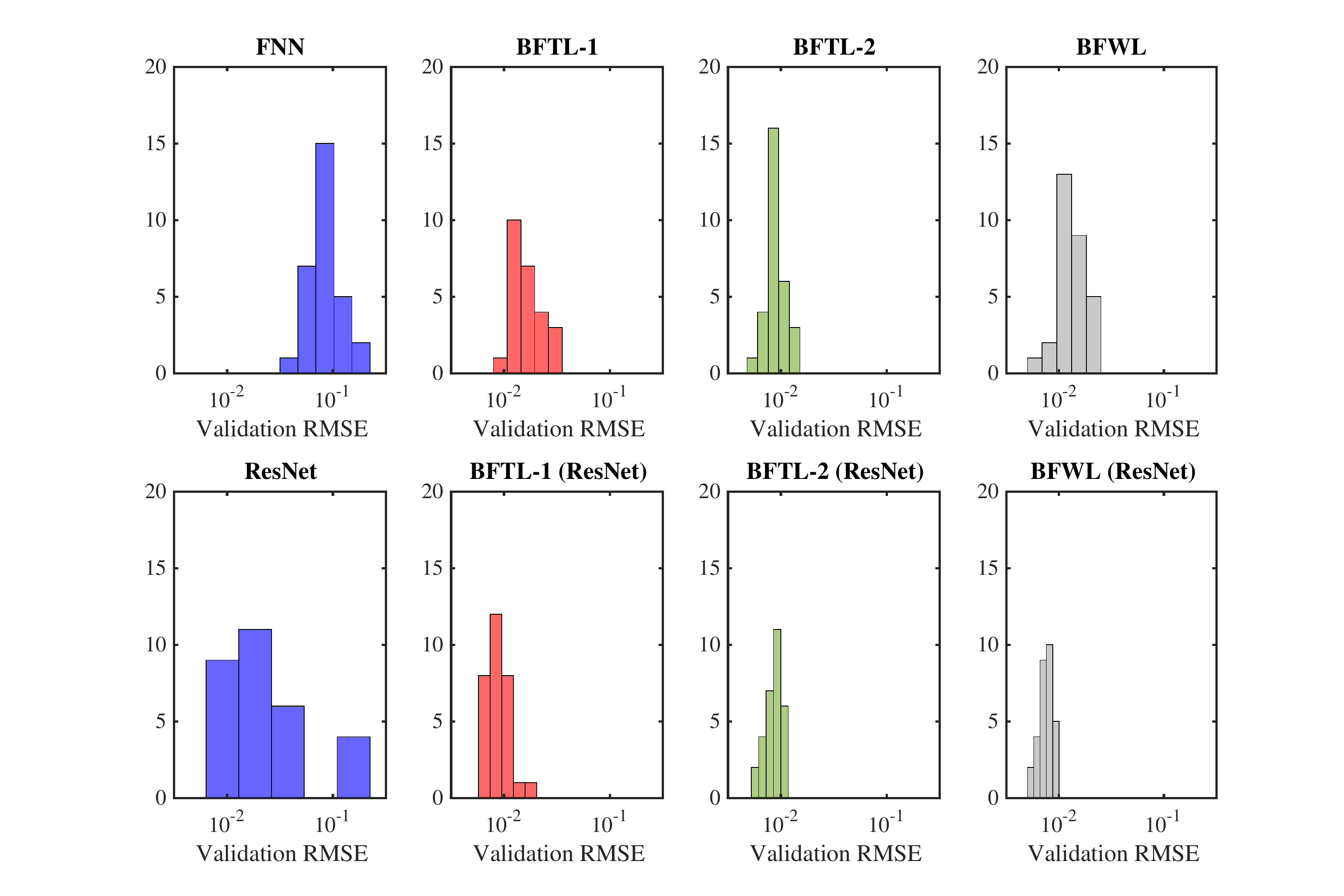}
    \caption{Histograms of the log of validation RMSEs for Example II using different learning approaches for 30 different validation datasets. The first column shows the validation RMSEs using standard learning techniques for two different architectures and $N_h=40$ high-fidelity samples. The rest of the columns show the validation RMSEs for different transfer learning strategies implemented using $N_h=20$ high-fidelity and $N_l=140$ low-fidelity samples.}
    \label{fig:battery_hist2}
\end{figure}

\subsubsection{Results}
\begin{table}[!htb]
 \caption{The learning rates used with different learning methods in Example II.}
    \label{tab:battery_lr}
    \centering
    \begin{tabular}{l|c|c|c}
    \hline \rule{0pt}{2ex} 
       Architecture & Method & Dataset & Learning rate, $\eta_k$\\
       \hline
\rule{0pt}{2ex}
\multirow{ 7}{*}{FNN} & Standard & $\Dh$ & $10^{-4}$\\\cline{2-4}
\rule{0pt}{2ex} & \multirow{2}{*}{BFTL-1} & $\Dl$ & $4\times10^{-4}$\\
& & $\Dh$ & $10^{-4}$\\\cline{2-4} \rule{0pt}{2ex}
& \multirow{2}{*}{BFTL-2} & $\Dl$ & $4\times10^{-4}$\\
& & $\Dh$ & $8\times10^{-3}$\\\cline{2-4} \rule{0pt}{2ex}
& \multirow{2}{*}{BFWL ($\pinn^\mathrm{s}$)} & $\Dl$ & $4\times10^{-4}$\\
& & $\Dh$ & $10^{-4}\quad (\beta = 0.01)$\\
\hline
\rule{0pt}{2ex}
\multirow{ 7}{*}{ResNet} & Standard & $\Dh$ & $4\times10^{-1}$\\\cline{2-4}
\rule{0pt}{2ex} & \multirow{2}{*}{BFTL-1} & $\Dl$ & $4\times10^{-4}$\\
& & $\Dh$ & $10^{-4}$\\\cline{2-4} \rule{0pt}{2ex}
& \multirow{2}{*}{BFTL-2} & $\Dl$ & $4\times10^{-4}$\\
& & $\Dh$ & $4\times10^{-3}$\\\cline{2-4} \rule{0pt}{2ex}
& \multirow{2}{*}{BFWL ($\pinn^\mathrm{s}$)} & $\Dl$ & $4\times10^{-4}$\\
& & $\Dh$ & $4\times10^{-3}\quad (\beta = 0.01)$\\ 
\hline
    \end{tabular}
\end{table}

\begin{table}[!htb]
 \caption{The mean validation RMSEs obtained using different learning methods and 30 different initializations of the network in Example II.}
    \label{tab:battery_mean_RMSE_I}
    \centering
    \begin{tabular}{l|c|c}
    \hline \rule{0pt}{2ex} 
       \multirow{2}{*}{Architecture} & \multirow{2}{*}{Method} & Mean validation\\
       & & RMSE\\
       \hline
\rule{0pt}{2ex}
\multirow{ 4}{*}{FNN} & Standard &  $9.3767\times10^{-2}$ \\
\rule{0pt}{2ex} & {BFTL-1} & $7.7482\times10^{-3}$ \\
 \rule{0pt}{2ex}
& {BFTL-2} & $8.1033\times10^{-3}$ \\
 \rule{0pt}{2ex}
& {BFWL} & $6.5568\times10^{-3}$\\
\hline
\rule{0pt}{2ex}
\multirow{ 4}{*}{ResNet} & Standard &  $1.1171\times10^{-2}$\\
\rule{0pt}{2ex} & {BFTL-1} & $6.9649\times10^{-3}$ \\
 \rule{0pt}{2ex}
& {BFTL-2} & $6.5039\times10^{-3}$ \\
 \rule{0pt}{2ex}
& {BFWL} & $7.0899\times10^{-3}$ \\
\hline
    \end{tabular}
\end{table}

\begin{table}[!htb]
 \caption{The mean validation RMSEs obtained using different learning methods and 30 random selections of training/validation datasets in Example II.}
    \label{tab:battery_mean_RMSE_II}
    \centering
    \begin{tabular}{l|c|c}
    \hline \rule{0pt}{2ex} 
       \multirow{2}{*}{Architecture} & \multirow{2}{*}{Method} & Mean validation\\
       & & RMSE\\
       \hline
\rule{0pt}{2ex}
\multirow{ 4}{*}{FNN} & Standard & $9.1903\times10^{-2}$ \\
\rule{0pt}{2ex} & {BFTL-1} & $1.7079\times10^{-2}$ \\
 \rule{0pt}{2ex}
& {BFTL-2} & $9.0712\times10^{-3}$ \\
 \rule{0pt}{2ex}
& {BFWL} & $1.3826\times10^{-2}$ \\
\hline
\rule{0pt}{2ex}
\multirow{ 4}{*}{ResNet} & Standard & $3.9497\times10^{-2}$ \\
\rule{0pt}{2ex} & {BFTL-1} & $8.5282\times10^{-3}$ \\
 \rule{0pt}{2ex}
& {BFTL-2} & $9.2004\times10^{-3}$ \\
 \rule{0pt}{2ex}
& {BFWL} & $7.6422\times10^{-3}$ \\
\hline
    \end{tabular}
\end{table}
We first use an FNN with two hidden layers and 50 neurons each using ELU activation functions with $\alpha=1$. We train this network with $N_h = 40$ high-fidelity samples. For the ResNet architecture, we add the output of the first hidden layer to the input of second hidden layer and use the ReLU activation functions for the second hidden layer. The transfer learning methodologies are implemented with the FNN and ResNet architectures using $N_l = 140$ and $N_h=20$ low- and high-fidelity samples, respectively. The Gaussian process teacher in BFWL uses a rational quadratic kernel with length scale bounds $(20,21)$ (see \ref{sec:gp}). The learning rates used with different methods are listed in Table \ref{tab:battery_lr}. 

Tables \ref{tab:battery_mean_RMSE_I} and \ref{tab:battery_mean_RMSE_II} show the mean validation RMSEs obtained from different methods for 30 independent random initializations of the network (with fixed training samples) and 30 random replications of the training/validation datasets (fixed network initialization), respectively. The samples used in the latter tests we selected randomly from the same pool of samples. Figs. \ref{fig:battery_hist1} and \ref{fig:battery_hist2} show the corresponding histograms. As suggested by Tables \ref{tab:battery_mean_RMSE_I} and \ref{tab:battery_mean_RMSE_II}, the three transfer learning techniques are able to reduce the mean validation error relative to the standard learning. These tables suggest also that using transfer learning the mean validation error does not change considerably when using ResNet instead of an FNN architecture. This shows that modeling the residual becomes less relevant as we use transfer learning. With standard learning, the ResNet architecture though provides smaller mean validation RMSEs compared to FNN, as illustrated in the first columns of Figs. \ref{fig:battery_hist1} and \ref{fig:battery_hist2}. 
\subsection{Example III: Flow around a NACA Airfoil}

In our last example, we consider flow around NACA 4412 airfoil at $Re=1.52\times10^6$ and low angle-of-attacks (AoA) as also studied in \cite{skinner2019reduced}. The airfoil has chord-length 1.0 m and is placed in a fluid domain of stream-wise length $L=999$ m ($x$-direction), vertical height $H=998$ m ($y$-direction), and span-wise width $W=2$ m ($z$-direction). This domain is shown in Fig. \ref{fig:ex3} along with the boundary conditions specified following the work of Diskin \textit{et al}. \cite{diskin2015grid}. The fluid enters the domain along the red line and exits at the blue lines in Fig. \ref{fig:ex3}. The boundaries at $\pm y$ and $\pm z$ directions are assumed to be inviscid and impenetrable. Further, no-slip boundary conditions on the airfoil surface are considered. To study the effect of the geometry variations, the maximum camber $m$, location of the maximum camber $p$,  maximum thickness as a fraction of the chord length $t$ are assumed to be parameters.  In addition the AoA $\alpha$ is also assumed to be a parameter. These parameters are modeled by uniform random variables and are specified in Table \ref{tab:airfoil_unc}.  Here, the QoI is the coefficient of pressure $C_p$ on the surface of the airfoil at a set of 153 grid points (every 5th surface grid point in the high-fidelity mesh shown in Fig. \ref{fig:airfoil_HF}). 

We use the mesh shown in Fig. \ref{fig:airfoil_LF} with 9,000 elements for the low-fidelity model to solve the Reynolds averaged Navier-Stokes (RANS) equations using PHASTA \cite{whiting2001stabilized}. We employ Spalart-Allmaras (SA) turbulence closure model \cite{spalart1992one,pope2001turbulent} in these simulations. The high-fidelity model uses a RANS simulation with SA closure and a mesh shown in Fig. \ref{fig:airfoil_HF}. This mesh is refined quadratically from near the airfoil to the boundaries and consists of 241,000 elements.  In the low-fidelity model, we do not resolve the shear layer downstream of the trailing edge in contrast to the high-fidelity model. The low-fidelity mesh is also unable to capture any wall effects and hence cannot produce a reasonably accurate solution. However, the computational cost of the low-fidelity model is 498 times smaller than that of the high-fidelity model. The interested readers is referred to \cite{skinner2019reduced} for more details of the airfoil model and its implementation.

\begin{figure}[!htb]
    \centering
    \includegraphics[scale=1.0]{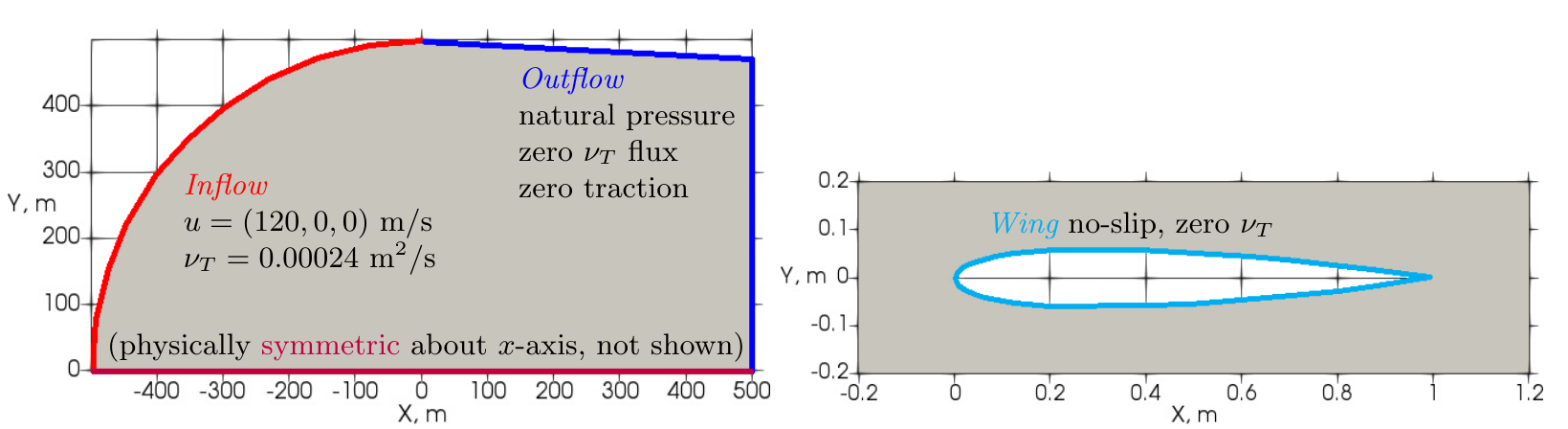}
    \caption{The schematic of the computational domain and the initial NACA 0012 airfoil used in Example III (adapted from \cite{skinner2019reduced}). We deform the mesh to map into NACA 4412 airfoil. Here, $u$ is the inflow velocity and $\nu_T$ is the kinematic turbulence viscosity.}
    \label{fig:ex3}
\end{figure}

\begin{table}[!htb]
 \caption{The uncertain parameters in Example III and their corresponding probability distributions. Note that $\mathcal{U}[a,b]$ denotes a uniform distribution between $a$ and $b$.}
    \label{tab:airfoil_unc}
    \centering
    \begin{tabular}{l|c}
    \hline \rule{0pt}{2ex} 
       Parameter  & Distribution \\
       \hline
\rule{0pt}{2ex}Maximum camber, $m$ &  $\mathcal{U}[0.032,0.048]$\\
        Location of maximum camber, $p$ &  $\mathcal{U}[0.32,0.48]$\\
        Maximum thickness, $t_{\max}$ &  $\mathcal{U}[0.096,0.144]$\\
        Angle of attack, $\alpha$ &  $\mathcal{U}[0^{\circ},6^{\circ}]$ \\\hline
    \end{tabular}
\end{table}

\begin{figure}[!htb]
    \centering
    \begin{subfigure}[t]{0.5\textwidth}
        \centering
        \includegraphics[scale = 1.2]{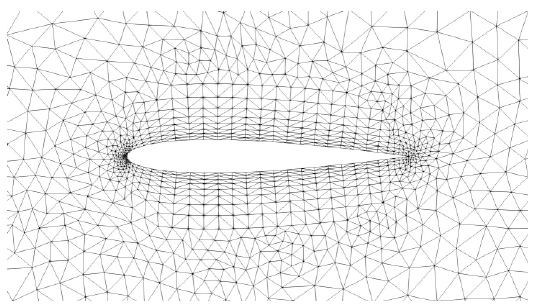}
        \caption{Low-fidelity mesh}\label{fig:airfoil_LF}
    \end{subfigure}%
    ~ 
    \begin{subfigure}[t]{0.5\textwidth}
        \centering
        \includegraphics[scale = 1.2]{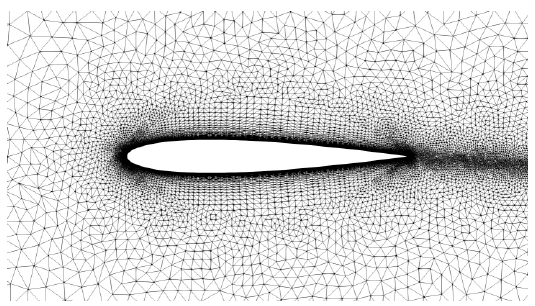}
        \caption{High-fidelity mesh}\label{fig:airfoil_HF}
    \end{subfigure}
    \caption{Meshes used for (a) low-fidelity model and (b) high-fidelity model in Example III for the NACA 0012 airfoil. They are then deformed to map into the NACA 4412 airfoil. Figure is adapted from \cite{skinner2019reduced}.}
\end{figure}\label{fig:ex3_mesh}

\subsubsection{Results} 

Four hidden layers with 50 neurons each and ELU activation functions is used for an FNN that is trained using $N_h=100$ high-fidelity samples for each of the 153 grid points. For validation, we use $N_v=100$ samples not used in the training. The ResNet architecture is implemented with the output from the first hidden layer added to the input of the fourth hidden layer. For the bi-fidelity trainings, we use $N_l =200$ low-fidelity samples along with $N_h=100$ high-fidelity samples. In BFTL-2, we add another hidden layer of 50 neurons to model the relation between $y_l$ and $y_h$. In the BFWL approach, we first train the network using $N_l=200$ low-fidelity samples and use $N_h=100$ high-fidelity samples to train the Gaussian process teacher with a Matern kernel (generalization of the RBF kernel) with length scale parameter $\rho=1.0$ and $\nu=2.5$ (see \ref{sec:gp}). Table \ref{tab:airfoil_lr} shows the learning rates used in different learning methods.

\begin{table}[!htb]
 \caption{The learning rates used with different learning methods in Example III.}
    \label{tab:airfoil_lr}
    \centering
    \begin{tabular}{l|c|c|c}
    \hline \rule{0pt}{2ex} 
       Architecture & Method & Dataset & Learning rate, $\eta_k$\\
       \hline
\rule{0pt}{2ex}
\multirow{ 7}{*}{FNN} & Standard & $\Dh$ & $4\times10^{-4}$\\\cline{2-4}
\rule{0pt}{2ex} & \multirow{2}{*}{BFTL-1} & $\Dl$ & $10^{-4}$\\
& & $\Dh$ & $10^{-4}$\\\cline{2-4} \rule{0pt}{2ex}
& \multirow{2}{*}{BFTL-2} & $\Dl$ & $4\times10^{-4}$\\
& & $\Dh$ & $10^{-3}$\\\cline{2-4} \rule{0pt}{2ex}
& \multirow{2}{*}{BFWL ($\pinn^\mathrm{s}$)} & $\Dl$ & $10^{-4}$\\
& & $\Dh$ & $10^{-4}\quad(\beta = 0.25)$\\
\hline
\rule{0pt}{2ex}
{ResNet} & Standard & $\Dh$ & $4\times10^{-4}$\\
\hline
    \end{tabular}
\end{table}

\begin{table}[!htb]
 \caption{The mean validation RMSEs obtained using different learning methods in Example III.}
    \label{tab:airfoil_mean_RMSE}
    \centering
    \begin{tabular}{l|c|c}
    \hline \rule{0pt}{2ex} 
       \multirow{2}{*}{Architecture} & \multirow{2}{*}{Method} & Mean validation\\
       & & RMSE\\
       \hline
\rule{0pt}{2ex}
\multirow{ 4}{*}{FNN} & Standard & $3.0076\times10^{-1}$ \\
\rule{0pt}{2ex} & {BFTL-1} & $3.1631\times10^{-1}$ \\
 \rule{0pt}{2ex}
& {BFTL-2} & $1.4896\times10^{-1}$ \\
 \rule{0pt}{2ex}
& {BFWL} & $3.2402\times10^{-2}$ \\
\hline
\rule{0pt}{2ex}
{ResNet} & Standard & $2.9170\times10^{-1}$ \\
\hline
    \end{tabular}
\end{table}

The histograms of the validation RMSEs for all 153 grid points using different learning methods are shown in Fig. \ref{fig:airfoil_hist}. 
Table \ref{tab:airfoil_mean_RMSE} shows the mean validation RMSEs obtained using different learning techiniques.
From this table, the mean error using the FNN is $3.01\times10^{-1}$ when trained only using the high-fidelity data. 
This does not improve significantly using the ResNet architecture (mean validation RMSE only becomes $2.9170\times10^{-1}$). Similar mean validation RMSEs have been observed for other combinations of the residual skip connections which we do not reported here. Due to these relatively large errors, we do not implement different transfer learning algorithms with the ResNet architecture. Among different transfer learning methods, the BFTL-2 approach slightly improves the validation errors as shown in Fig. \ref{fig:airfoil_hist} and Table \ref{tab:airfoil_mean_RMSE}. 
On the other hand, the mean validation RMSE using the BFWL approach is $3.2402\times10^{-2}$, which shows an order of magnitude improvement compared to all other learning methods. This again shows the usefulness of a Gaussian teacher and the weighted learning rate in fine-tuning the low-fidelity student network. 

\begin{figure}[!htb]
    \centering
    \includegraphics[scale = 0.55]{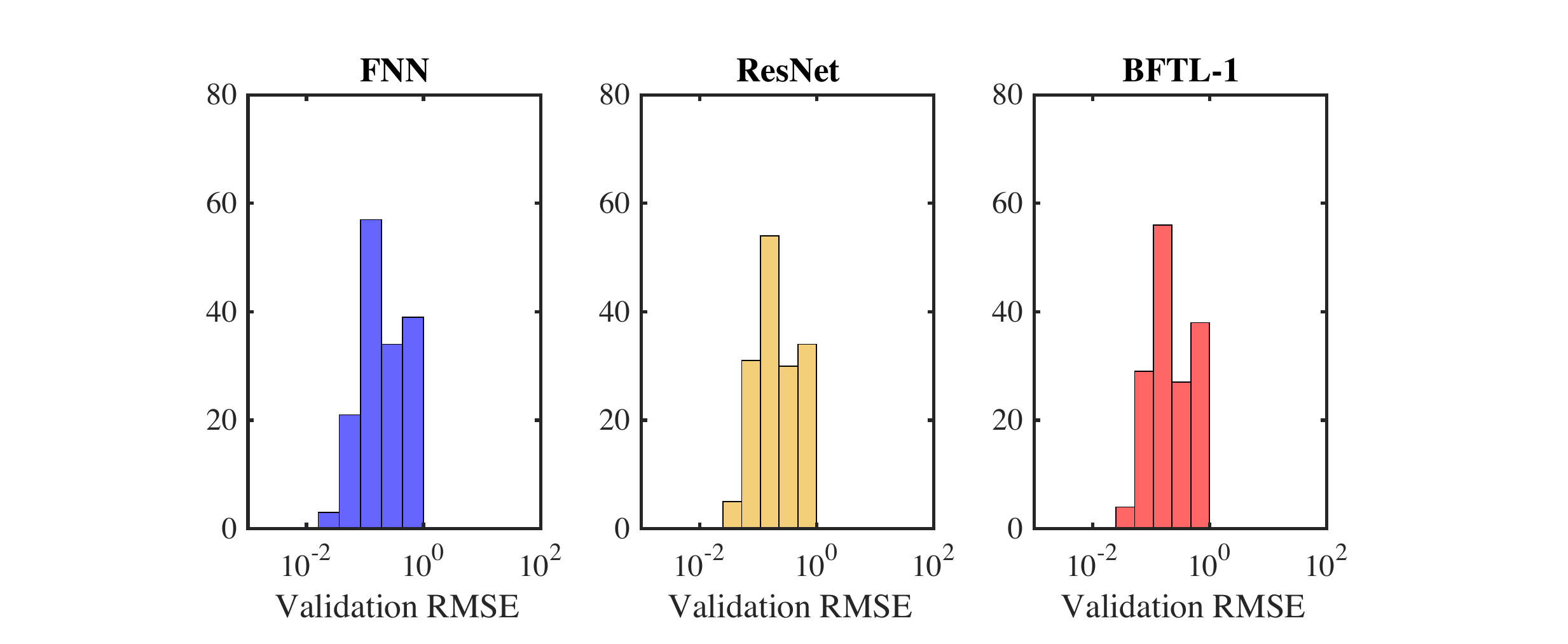}\\
    \includegraphics[scale = 0.55]{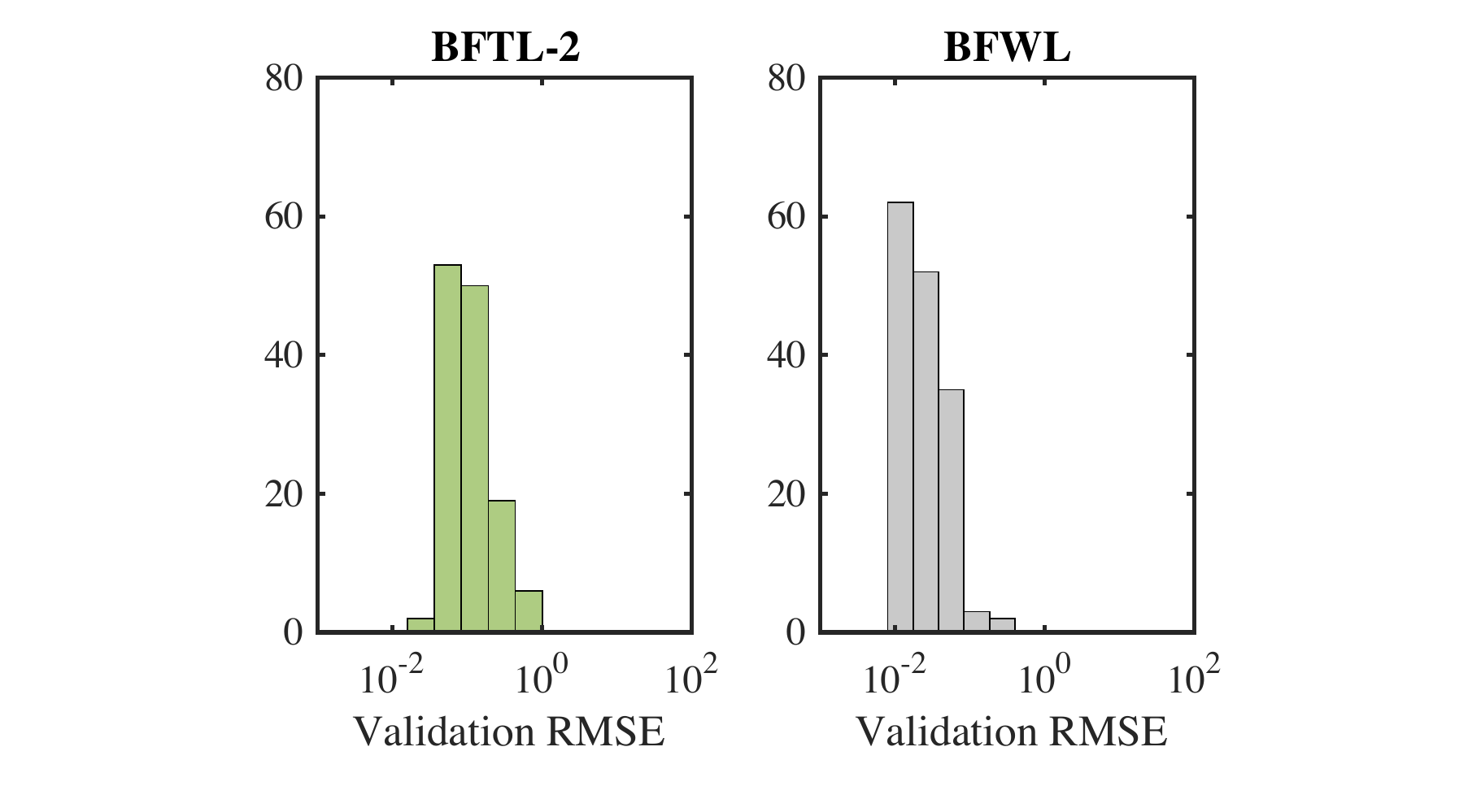}
    \caption{Histograms of the validation RMSEs in the estimation of $C_p$ for Example III using different learning approaches for 153 grid points along the surface of the airfoil. 
    The first two histograms show the errors using the standard learning techniques for two different architectures and $N_h=100$ high-fidelity samples. The rest of the histograms show validation RMSEs for different transfer learning strategies implemented using $N_h=100$ high-fidelity and $N_l=200$ low-fidelity samples.}
    \label{fig:airfoil_hist}
\end{figure}

\section{Conclusions}

Recently, there has been an increased interest in using neural networks for modeling uncertainty propagation through complex engineering systems. With computational resources specifically built for training these networks, it has become more relevant to exploit the expressiveness of neural networks for UQ. However, for large-scale engineering and scientific applications the main obstacle in training such networks is generating large datasets either using high-fidelity models or costly experimental setups. With the goal of reducing the amount of required data for training, transfer learning employs an already trained network for similar applications and adapts it to the application at hand.

In this paper, we study the utility of transfer learning using bi-fidelity data, where we train the network using low-fidelity data and adapt it using a smaller high-fidelity dataset. In our first implementation (BFTL-1), the uppermost hidden layer of a network trained using the low-fidelity dataset is fine-tuned using the high-fidelity dataset. In a second approach (BFTL-2), we add a shallow network to model the map between the QoI estimated by the low-fidelity and its high-fidelity counterpart. In our third implementation (BFWL), we modify the parameters of the low-fidelity neural network with synthetic data (with the associated uncertainty) generated via a high-fidelity trained Gaussian process. We explore these transfer learning techniques using three numerical examples. The first numerical example from structural mechanics shows that transfer learning reduces the validation error, especially as compared to the standard feed-forward network. The largest reduction -- more than an order of magnitude -- is achieved by BFWL. A similar observation is made from the second example on a Lithium ion battery model involving non-linear, multi-scale, and multi-physics phenomena subject to high-dimensional uncertainty. In the third example involving an airfoil with parameterized geometry and inflow angle-of-attack, we again observe the efficacy of BFWL, while the first two approaches lead to accuracies similar to that of the standard training. From these examples, we observe BFWL performs the best among the considered transfer learning strategies and leads to considerable error improvements over standard training. Additionally, the reduction in the error achieved by the transfer learning techniques depends on the architecture of the network, here feed-forward and residual (ResNet) networks.

We expect transfer learning to be an important tool for understanding complex physics and the effects of uncertainty in the near future. Initial works, such as the techniques communicated in this paper, show promise producing accurate surrogate models capable of rapidly generating higher-fidelity samples.  However, there is still a lot of work required for these tools to reach the level of maturity present in other widely-used multi-fidelity  techniques discussed in Section \ref{sec:intro}. 

\section{Acknowledgment}
This work was authored in part by the National Renewable Energy Laboratory, operated by Alliance for Sustainable Energy, LLC, for the U.S. Department of Energy (DOE) under Contract No. DE-AC36-08GO28308. This material is based upon work supported by NSF grants CMMI-1454601 and OAC-1740330, as well as the DARPA grant HR0011-17-2-0022. The views expressed in the article do not necessarily represent the views of the DOE or the U.S. Government. The U.S. Government retains and the publisher, by accepting the article for publication, acknowledges that the U.S. Government retains a nonexclusive, paid-up, irrevocable, worldwide license to publish or reproduce the published form of this work, or allow others to do so, for U.S. Government purposes.







\appendix
\section{Gaussian Process Regression or Kriging} \label{sec:gp}
A zero-mean Gaussian process $f(\xm) \sim \mathcal{GP}\Big(0, \kappa(\xm,\xm^{\prime}) \Big)$ with covariance function $\kappa(\xm,\xm^{\prime})$ can be written in terms of the following Gaussian distribution \cite{rasmussen2004gaussian}
\begin{equation}\left\{
    \begin{array}{c}
         f(\xm)  \\
         f(\xm^\prime) \\
    \end{array} \right\} \sim \mathcal{N}\left( \mathbf{0}, \left[ 
    \begin{array}{cc}
        \kappa(\xm,\xm) & \kappa(\xm,\xm^\prime) \\
        \kappa(\xm^\prime,\xm) & \kappa(\xm^\prime,\xm^\prime)\\
    \end{array}
    \right]\right).
\end{equation}
In the presence of independent zero-mean Gaussian noise $\varepsilon$ with variance $\sigma_n^2$ and measured data $\ym=(y_1,\dots,y_N)$, where
\begin{equation} 
\label{eqn:noisy_samples}
y_i  = f(\xm_i) +\varepsilon_i,
\end{equation}
the joint probability density for prediction at $\xm^\prime$ become

\begin{equation}
\begin{split}
&       \left\{
    \begin{array}{c}
         \ym  \\
         f(\xm^\prime) \\
    \end{array} \right\} \sim \mathcal{N}\left( \mathbf{0}, \left[ 
    \begin{array}{cc}
        \Km+\sigma_n^2\mathbf{I} & \km_{f^\prime} \\
        \km^{T}_{f^\prime} & \kappa_{f^\prime f^\prime}\\
    \end{array}
    \right]\right).
\end{split}
\end{equation}
Here, $\mathbf{I}$ is the identity matrix, $\Km$ is the $N\times N$ covariance matrix associated with the inputs $\Xm := \{\xm_i\}_{i=1}^N$ such that $\Km(i,j)=\kappa(\xm_i,\xm_j)$, and $\km_{f^\prime}$ is the $N\times 1$ covariance vector between all the inputs $\Xm$ and $\xm^\prime$, i.e., $\km_{f^\prime}(i) = \kappa(\xm_i,\xm^\prime)$, and $\kappa_{f^\prime f^\prime} = \kappa(\xm^\prime,\xm^\prime)$. Hence, the posterior density of $f(\xm^\prime)$ is given by
\begin{equation}
    f(\xm^\prime)\big\lvert \ym\sim \mathcal{N}\left( \bar{f}^\prime, \sigma^2_{f^\prime} \right),
\end{equation}
\sloppy where $\bar{f}^\prime=\km_{f^\prime}^{T}\left[ \Km+\sigma_n^2\mathbf{I} \right]^{-1}\ym$ and $\sigma_{f^\prime}^2 = \kappa_{f^\prime f^\prime} - \km_{f^\prime}^{T}\left[ \Km+\sigma_n^2\mathbf{I} \right]^{-1}\km_{f^\prime}$. 
Note that many choices are available for the covariance function, \textit{e.g.}, radial basis kernel $\kappa(\xm, \xm^\prime;\alpha,\sigma) = \alpha^2\exp\left(-\frac{\|\xm - \xm^\prime\|^2}{2\sigma^2}\right)$, rational quadratic kernel $\kappa(\xm,\xm^\prime;\alpha,\rho,\sigma)=\alpha^2\exp\left(1+\frac{\|\xm - \xm^\prime\|^2}{2\rho\sigma^2}\right)^{-\rho}$, Matern kernel $\kappa(\xm,\xm^\prime;\alpha,\rho,\nu) = \alpha^2\frac{2^{1-\nu}}{\Gamma(\nu)}\left(\frac{\sqrt{2\nu}\lVert\xm-\xm^\prime\rVert}{\rho}\right)^\nu B_{\nu}\left(\frac{\sqrt{2\nu}\lVert\xm-\xm^\prime\rVert}{\rho}\right)$, white noise kernel $\kappa(\xm,\xm^\prime;\alpha)=\alpha^2\delta_{\xm,\xm^\prime}$, etc., where $\alpha$, $\rho$, $\sigma$, and $\nu$ are hyperparameters; $\Gamma(\cdot)$ is the Gamma function; $B_\nu(\cdot)$ is the modified Bessel function of the second kind; and $\delta_{\cdot,\cdot}$ is the Kronecker delta function. 
These hyperparameters of the covariance function and the noise variance $\sigma_n^2$ (collectively referred to as $\bm\theta$) are inferred using the measurement data by maximizing the likelihood or minimizing the following negative log-likelihood
\begin{equation}
    -\log{\mathrm{p}(\ym|\bm\theta)} = \frac{1}{2}\ym^T\left( \Km+\sigma_n^2\mathbf{I} \right)^{-1} \ym +\frac{1}{2}\log \lvert\Km+\sigma_n^2\mathbf{I}\rvert+\frac{N}{2}\log(2\pi).
\end{equation}

\section{Multi-fidelity Gaussian Process Regression or Co-kriging}\label{sec:cokriging}

Using the low-fidelity dataset $\D_{l}=\{\xm_{l,i},y_{l,i}\}_{i=1}^{N_l}$ and a (smaller) high-fidelity dataset $\D_{h}=\{\xm_{h,i},y_{h,i}\}_{i=1}^{N_h}$, co-kriging {\cite{Kennedy00,Qian06}} build the following multiplicative-additive relation between the low- and high-fidelity models,
\begin{equation}\label{eq:co-kriging}
    f_{h}(\xm) = \rho f_{l}(\xm) + f_{\mathrm{corr}}(\xm).
\end{equation}
Here, $f_{l}(\xm)\sim\mathcal{GP}(0,\kappa_{l}\left(\xm,\xm^\prime)\right)$ is the Gaussian process approximation of the low-fidelity data (here for simplicity assumed to be of zero mean). The additive correction term $f_{\mathrm{corr}}(\xm)\sim\mathcal{GP}(0,\kappa_{\mathrm{corr}}\left(\xm,\xm^\prime)\right)$ is also a Gaussian process independent of $f_{l}(\xm)$, and $\rho$ is a multiplicative scaling factor (but can also be a function of $\xm$). Following (\ref{eq:co-kriging}),  $f_{h}(\xm)$ is also a zero-mean Gaussian process with covariance function $\rho^2\kappa_{l}\left(\xm,\xm^\prime\right)+\kappa_{\mathrm{corr}}\left(\xm,\xm^\prime\right)$. Furthermore, we assume $\ym_{h} = (y_{h,1},\dots,y_{h,N})$ and $\ym_{l}= (y_{l,1},\dots,y_{l,N})$ are generated in a manner similar to (\ref{eqn:noisy_samples}) from their respective models and with independent, zero-mean additive Gaussian noises with variances $\sigma_{h}^2$ and $\sigma_{l}^2$, respectively. Hence, the joint probability density for the prediction at $\xm^\prime$ becomes
\begin{equation}
\begin{split}
&\left\{
    \begin{array}{c}
         \ym_{h} \\
         \ym_{l}\\
         f(\xm^\prime) \\
    \end{array} \right\} \sim \mathcal{N}\left( \mathbf{0}, \left[
    \begin{array}{ccc}
    \Km_{hh} & \Km_{hl} & \km_{{h}f^\prime} \\
    \Km_{lh}  & \Km_{ll} & \km_{{l}f^\prime}\\
    \km_{f^\prime{h}} & \km_{f^\prime\mathrm{l}} & \kappa_{f^\prime f^\prime}\\
    \end{array}
    \right]\right),\\
    \end{split}
\end{equation}
where $\Km_{hh}(i,j) = \rho^2\kappa_l(\xm_{h,i},\xm_{h,j})+\kappa_\mathrm{corr}(\xm_{h,i},\xm_{h,i})+\sigma_h^2\delta_{i,j}$, $\Km_{hl}(i,j) = \rho\kappa_l(\xm_{h,i},\xm_{l,j})$, $\km_{{h}f^\prime}(i)=\rho^2\kappa_l(\xm_{h,i},\xm^\prime)+\kappa_\mathrm{corr}(\xm_{h,i},\xm^\prime)$, $\Km_{ll}(i,j)=\kappa_l(\xm_{l,i},\xm_{j,l})+\sigma_l^2\delta_{i,j}$, $\km_{{l}f^\prime}(i)=\rho\kappa_l(\xm_{l,i},\xm^\prime)$, and $\kappa_{f^\prime f^\prime}=\rho^2\kappa_l(\xm^\prime,\xm^\prime)+\kappa_\mathrm{corr}(\xm^\prime,\xm^\prime)$, $\Km_{lh}=\Km^{T}_{lh}$, $\km_{f^\prime h}=\km^{T}_{h f^\prime}$, and $\km_{f^\prime l}=\km^{T}_{l f^\prime}$.
The predictive density of $f(\xm^\prime)$ is given by
\begin{equation}
    f(\xm^\prime)\big\lvert \ym_l,\ym_h \sim \mathcal{N}\left( \bar{f}^\prime, \sigma^2_{f^\prime} \right),
\end{equation}
where $\bar{f}^\prime=\left\{\begin{array}{c}
    \km_{{h}f^\prime} \\ \km_{{l}f^\prime}  \\
\end{array}\right\}^T\left[
    \begin{array}{ccc}
    \Km_{hh} & \Km_{hl} \\
    \Km_{lh} & \Km_{ll} \\
    \end{array}
    \right]^{-1}\left\{
    \begin{array}{c}
         \ym_{h}  \\
         \ym_{l}\\
    \end{array} \right\}$ and $\sigma_{f^\prime}^2 = \kappa_{f^\prime f^\prime} - \left\{\begin{array}{c}
    \km_{{h}f^\prime} \\ \km_{{l}f^\prime}  \\
\end{array}\right\}^T\left[
    \begin{array}{ccc}
    \Km_{hh} & \Km_{hl} \\
    \Km_{lh} & \Km_{ll} \\
    \end{array}
    \right]^{-1}\left\{\begin{array}{c}
    \km_{{h}f^\prime} \\ \km_{{l}f^\prime}  \\
\end{array}\right\}$. 
The hyperparameters $\bm\theta$ of the covariance functions and the noise variances are inferred using the measurement data by maximizing the likelihood or minimizing the following negative log-likelihood
\begin{equation}
\begin{split}
    -\log{\mathrm{p}(\ym_h,\ym_l |\bm\theta)} =& \frac{1}{2}\left\{
    \begin{array}{c}
         \ym_{h}  \\
         \ym_{l}\\
    \end{array} \right\}^T\left[
    \begin{array}{ccc}
    \Km_{hh} & \Km_{hl} \\
    \Km_{lh} & \Km_{ll} \\
    \end{array}
    \right]^{-1}\left\{
    \begin{array}{c}
         \ym_{h}  \\
         \ym_{l}\\
    \end{array} \right\}\\
    &+\frac{1}{2}\log \Bigg\lvert \left[
    \begin{array}{ccc} 
    \Km_{hh} & \Km_{hl} \\
    \Km_{lh} & \Km_{ll} \\
    \end{array}\right]
    \Bigg\rvert+\frac{N_l+N_h}{2}\log(2\pi).
    \end{split}
\end{equation}

\section{Uncertain Parameters in Example II} \label{sec:exIIparam}

The 17 uncertain parameters used in Section \ref{sec:exII} are described below in brief.

\begin{itemize}
\item \textit{Porosity, $\epsilon$}: Porosity is defined as the ratio of the voids to the total volume. In a Li-ion battery, we have $\epsilon_\mathrm{a}$, $\epsilon_\mathrm{c}$, and $\epsilon_\mathrm{s}$ --- porosity in the anode, cathode, and separator, respectively. The power generated by the battery is directly proportional but the capacity of the battery is inversely proportional to the porosity.

\item\textit{Bruggeman coefficient, $b$}:
The relation between the tortuosity and the porosity is given by the Bruggeman relation,
\begin{equation}
    \tau = \epsilon^{(1-b)},
\end{equation}
where \textit{tortuosity} $\tau$ is a geometric parameter of the electrodes that affects the effective transport properties and $b$ is the Bruggeman coefficient.
Here, we have three Bruggeman coefficients, $b_\mathrm{a}$, $b_\mathrm{c}$, and $b_\mathrm{s}$, for the anode, cathode, and separator, respectively.

\item\textit{Solid diffusion coefficient, $D_\mathrm{s}$}:
These coefficients, $D_\mathrm{s,a}$ and $D_\mathrm{s,c}$, determine the diffusivity of the ions from the cathode and anode, respectively.

\item\textit{Solid conductivity, $\sigma$}:
The flow of the current through the anode and cathode is characterized by the conductivities $\sigma_\mathrm{a}$ and $\sigma_\mathrm{c}$, respectively.

\item\textit{Reaction rate, $k$}:
The rate of the chemical reaction in the anode and cathode are described by the reaction rates $k_\mathrm{a}$ and $k_\mathrm{c}$, respectively.

\item\textit{Component length, $L$}:
The length of the anode, cathode, and separator --- $L_\mathrm{a}$, $L_\mathrm{c}$, and $L_\mathrm{s}$, respectively --- are assumed to be uncertain in Example II in Section \ref{sec:exII}.

\item\textit{Salt diffusion coefficient, $D$}:
The friction between the salt ions (formed during the polarization) and the solvent can adversely affect the battery power. Hence, a larger value of salt diffusion coefficient $D$ is preferred.

\item\textit{Li$^+$ transference number, $t_+^0$}: This number denotes the amount of current carried through the Li$^+$ ions.
\end{itemize}

\bibliographystyle{IJ4UQ_Bibliography_Style}

\bibliography{references}
\end{document}